\documentclass{article}

\usepackage[preprint]{neurips_2025}

\usepackage[utf8]{inputenc} % allow utf-8 input
\usepackage[T1]{fontenc}    % use 8-bit T1 fonts
\usepackage{hyperref}       % hyperlinks
\usepackage{url}            % simple URL typesetting
\usepackage{booktabs}       % professional-quality tables
\usepackage{amsfonts}       % blackboard math symbols
\usepackage{nicefrac}       % compact symbols for 1/2, etc.
\usepackage{microtype}      % microtypography
\usepackage{xcolor}         % colors
\usepackage{amsmath}
\usepackage{amssymb}
\usepackage{mathtools}
\usepackage{amsthm}
\usepackage{multirow}
\usepackage{algorithm}
\usepackage{algorithmic}
\usepackage{amsmath,amsfonts,bm}

\def\eqref#1{equation~\ref{#1}}
\def\1{\bm{1}}

\def\va{{\bm{a}}}
\def\vb{{\bm{b}}}

\def\vw{{\bm{w}}}

\def\vz{{\bm{z}}}

\def\mW{{\bm{W}}}

\DeclareMathAlphabet{\mathsfit}{\encodingdefault}{\sfdefault}{m}{sl}
\SetMathAlphabet{\mathsfit}{bold}{\encodingdefault}{\sfdefault}{bx}{n}

\def\gE{{\mathcal{E}}}
\def\gF{{\mathcal{F}}}

\def\gO{{\mathcal{O}}}

\def\gR{{\mathcal{R}}}

\def\gX{{\mathcal{X}}}
\def\gY{{\mathcal{Y}}}

\def\sN{{\mathbb{N}}}

\def\sR{{\mathbb{R}}}

\newcommand{\E}{\mathbb{E}}

\DeclareMathOperator*{\argmin}{arg\,min}

\makeatletter
\newtheorem{rep@theorem}{\rep@title}
\newcommand{\newreptheorem}[2]{%
  \newenvironment{rep#1}[1]{%
    \def\rep@title{#2 \ref{##1}}%
    \begin{rep@theorem}}%
    {\end{rep@theorem}}}
\makeatother

\theoremstyle{plain}
\newtheorem{theorem}{Theorem}[section]
\newreptheorem{theorem}{Theorem} 

\newtheorem{lemma}[theorem]{Lemma}

\theoremstyle{definition}
\newtheorem{definition}[theorem]{Definition}
\newtheorem{assumption}[theorem]{Assumption}
\theoremstyle{remark}

\title{Federated Adapter on Foundation Models:  \\An Out-Of-Distribution Approach}

\author{%
  Yiyuan Yang \\
  University of Technology Sydney\\
  \texttt{yiyuan.yang-1@student.uts.edu.au} \\
  \And
  Guodong Long\\
  University of Technology Sydney\\
  \texttt{Guodong.Long@uts.edu.au} \\
  \And
  Tianyi Zhou \\
  University of Maryland \\
  \texttt{zhou@umiacs.umd.edu} \\
  \And
  Qinghua Lu\\
  Data61, CSIRO \\
  \texttt{qinghua.lu@data61.csiro.au} \\
  \And
  Shanshan Ye \&  Jing Jiang\\
  University of Technology Sydney \\
  \texttt{\{Shanshan.Ye, Jing.Jiang\}@uts.edu.au} \\
}

\begin{document}

\maketitle

\begin{abstract}
  As foundation models gain prominence, Federated Foundation Models (FedFM) have emerged as a privacy-preserving approach to collaboratively fine-tune models in federated learning (FL) frameworks using distributed datasets across clients. A key challenge for FedFM, given the versatile nature of foundation models, is addressing out-of-distribution (OOD) generalization, where unseen tasks or clients may exhibit distribution shifts leading to suboptimal performance. 
Although numerous studies have explored OOD generalization in conventional FL, these methods are inadequate for FedFM due to the challenges posed by large parameter scales and increased data heterogeneity. 
To address these, we propose FedOA, which employs adapter-based parameter-efficient fine-tuning methods for efficacy and introduces personalized adapters with feature distance-based regularization to align distributions and guarantee OOD generalization for each client.
Theoretically, we demonstrate that the conventional aggregated global model in FedFM inherently retains OOD generalization capabilities, and our proposed method enhances the personalized model's OOD generalization through regularization informed by the global model, with proven convergence under general non-convex settings.
Empirically, the effectiveness of the proposed method is validated on benchmark datasets across various NLP tasks.
\end{abstract}

\section{Introduction}
% FedFM
Recently, Foundation models have gained significant attention for their versatility in handling diverse downstream tasks. However, their reliance on large volumes of public data raises challenges as data resources become scarce. To address this, Federated Foundation Models (FedFM) \cite{zhuang2023foundation,yu2023federated} have been proposed as a promising solution by leveraging federated learning (FL) to enable distributed training across devices or data sources while keeping private data localized and secure.  

% OOD (in-distribution generalization)
Out-of-distribution (OOD) generalization constitutes a pivotal research challenge that aims to train models capable of performing robustly on data exhibiting distributions different from those seen during training. This challenge has been extensively explored across various centralized research areas \cite{liu2021towards,arjovsky2020out}, and recent scholarly efforts have extended these methodologies to federated learning frameworks \cite{li2023federated,yuan2021we}, where some unseen (non-participation during training) tasks/clients may exhibit distribution shifts leading to suboptimal performance of the conventional FL methods. 

\textbf{Heterogeneity in OOD and FL.}
Data heterogeneity presents a significant challenge in both OOD and FL. The key difference between FL and OOD lies in the sources of heterogeneity with their respective evaluation data. In FL, data heterogeneity primarily arises from various training clients, with a focus on in-distribution performance, where test data are from the same clients as the training data, reflecting a composite of the training environments.
In contrast, OOD generalization addresses heterogeneity arising from distribution shifts between training and testing data, emphasizing performance on diverse, unseen distributions to test broader generalization capabilities.
Therefore, unlike prior FL research \cite{chen2021bridging,zhou2023fedfa,xie2024perada} focused on in-distribution generalization by evaluating client performance within training environments, OOD generalization in FL requires methods that address distribution shifts both among clients and between training and testing data to ensure robust performance.
% In OOD, data heterogeneity present within the training dataset is centralized for model learning. In FL, data heterogeneity manifests as statistical variations across numerous clients, each contributing uniquely distributed data. 
% The key difference between FL and OOD lies in their respective evaluation methodologies. 
% In FL, the evaluation typically involves test data sourced from the same clients that provide the training data, essentially reflecting a composite of the training environments. In contrast, OOD generalization contends with a broader spectrum of test data diversity, where the distributions may significantly deviate from those encountered during training, thus posing a more stringent test of the model’s generalization capabilities. Therefore, different from the 
% Therefore, conventional FL methods cannot perform well in OOD scenrios, which prompts adapting OOD algorithms in FL for better generalization.

% motivation 
Although numerous approaches \cite{guo2023out,tang2023learning} have been proposed to address OOD generalization in conventional FL, they may not be optimal for FedFM. A key challenge in FedFM arises from the \emph{large parameter scale} of foundation models \cite{ren2024advances}. Unlike conventional FL primarily focuses on smaller models, FedFM typically utilizes foundation models with billions of parameters, leading to substantial communication and computation costs when operating on the entire model. To mitigate these issues, recent research \cite{kuang2023federatedscope,zhang2023fedpetuning} in FedFM has adapted parameter-efficient fine-tuning (PEFT) methods, where only a small subset of parameters is learned and communicated for efficacy. However, simply adapting conventional OOD FL methods to PEFT-based FedFM would suffer from \emph{structural heterogeneity} \cite{sun2024improving}, particularly in adapter-based methods \cite{hu2023llm}, where joint optimization conflicts with the separate operation of adapter parameters, undermining performance. Another significant challenge for FedFM is the \emph{increased data heterogeneity}, such as cross-domain data, due to the versatile nature of foundation models, which are designed to handle a variety of downstream tasks in real-world applications \cite{liu2023summary}. Therefore, it is crucial to explore innovative approaches to address these challenges for effective OOD generalization in FedFM. 

Previous work \cite{du2024risk} first analyzed the OOD generalization of FedFM through robustness experiments and proposed a noisy projection-based robust aggregation algorithm, but still rooted in the conventional non-IID (heterogeneous label distributions) setting of FL, overlooks adapter structural heterogeneity, and lacks comprehensive theoretical analysis. 
To fill these gaps, we propose FedOA, a novel framework that adapts invariant learning \cite{arjovsky2019invariant,koyama2020invariance}—a widely used approach for centralized OOD that learns invariant features consistent across distributions—for OOD generalization in FedFM.
% while addressing the substantial communication and computation costs associated with more heterogeneous scenarios. Our approach begins by revisiting existing invariant learning techniques in conventional FL, reformulating them into a unified optimization framework. 
We first theoretically analyze the generalization bounds of both the conventional aggregated global model and the personalized model in FedFM, demonstrating that the global model inherently retains OOD generalization ability. This motivates our approach to enhance the personalized model's OOD generalization by leveraging the global model. Specifically, we employ adapter-based PEFT methods for efficient learning and incorporate personalized adapters to address client-specific needs. Additionally, we introduce a feature distance-based regularization term to improve OOD generalization of personalized adapter by learning from the global model and mitigating structural heterogeneity in PEFT methods. Finally, we provide a theoretical framework to analyze the convergence of our method in FedFM. 
Our contributions are summarized below.

\begin{itemize} 

\item We introduce a new method, namely FedOA, to learn invariant features for addressing the OOD generalization of FedFM with large parameter scales in increased data heterogeneity scenarios.

\item We theoretically demonstrate that the conventional aggregated global model in FedFM inherently retains OOD generalization ability, and FedOA is expected to enhance OOD generalization through feature distance-based regularization. We also present the convergence results for FedOA under general non-convex settings.

\item We evaluate our method on heterogeneous FedFM benchmarks across diverse NLP tasks, demonstrating state-of-the-art performance and superior OOD generalization compared to existing methods.%We conduct an experimental analysis using heterogeneous FedFM benchmarks across diverse NLP tasks. Empirical outcomes reveal that our method attains state-of-the-art performance, underscoring its superior OOD generalization capabilities than existing methods.
\end{itemize}

% \section{Related Work}

\section{Preliminaries and Challenges}

\begin{table*}[t]
\caption{Table of partial notations.}
\label{table1}
\begin{center}
\begin{tabular}{l|l|l}
\toprule
\bf Components & \bf Notation & \bf Definition \\
\midrule
\multirow{6}{*}{OOD} & $(X, Y)$ & Random variables of inputs and outputs\\
& $f_\theta$& Hypothesis with parameter $\theta$\\
& $\ell(f(X),Y)$ & Loss function\\
& $(X^e, Y^e) \sim P_e$ & Probability distribution of environment $e$\\
& $\gE$ & Collection of environments $e$ \\
%& $\{ \sP^e \} _{e\in \gE}$ & Collection of probability distributions of environments $\gE$  \\
& $\gR(f) = \E_{(X,Y)\sim P}[\ell(f(X),Y)]$ & Expected risk of model $f$ \\
\midrule
%& $\gC$& Collection of clients \\
%& $m$ & Number of clients in FL system \\
\multirow{4}{*}{FL} & $S_e,|S_e|$ & The dataset and its size on Client $e$\\
& $\xi \sim S$ & Batch of samples from dataset $S$\\
& $K$ & Number of local update steps\\
& $T$ & Number of communication rounds\\
& $\eta_l,\eta_g$ & Local and global learning rates\\
& $ R(f) = \frac{1}{|S|}\sum_{(x_i,y_i)\in S}\ell(f(x_i),y_i)$ &  Empirical risk of model $f$ over data $S$ \\
\bottomrule
\end{tabular}
\end{center}
\end{table*}

\subsection{Preliminaries}

Let $\gX$ denote the feature space and $\gY$ the label space. There are often families of probability distributions $\{P_e\}_{e\in\gE}$ over the space $\gX \times \gY$, where the indices $e\in\gE$ represent different environments (also referred as ``domains''). Each distribution $P_e$ can be denoted as $(X^e, Y^e)\sim P_e$. $\gE_{all}$ is the collection of all possible environments, with $\gE_{train},\gE_{test} \subseteq \gE_{all}$ as training and testing environments respectively. The notations related to OOD generalization are delineated in the first part of Table~\ref{table1}, whereas the latter part elucidates components relevant to federated learning.

\paragraph{The Objective of OOD Generalization.}
In practical settings, there is often such a case in which test data originate from distributions that differ from those of the training data. OOD generalization is a research domain that specifically addresses these discrepancies. Following the conventional methodologies \cite{arjovsky2020out}, we assume that the distribution of the test data belongs to $\gE_{all}$ and the objective of OOD generalization is to minimize the worst case over all potential test distributions, which can be formulated as:
\begin{equation}
\label{ood_obj}
    \min_f \max_{e\in \gE_{all}} \gR_e(f),
\end{equation}
where $\gR_e(f) = \E_{(X^e,Y^e)\sim P_e}[\ell(f(X^e),Y^e)]$, $f$ is the model and $\ell$ is the loss function.

\paragraph{OOD Generalization in FL.}
In FL, the task in each client can be taken as an environment $e$ with a local dataset $S_e$ drawn from distribution $P_e$. Consequently, tasks in training clients can be taken as the collection of $\gE_{train}$, and $\gE_{all}$ represents all possible tasks/clients. The objective of OOD generalization in FL, therefore, aligns with the general objective in equation (\ref{ood_obj}). Specifically, due to the distributed nature of FL, OOD scenarios can occur within individual clients (\textbf{intra-client}) or across different clients (\textbf{inter-client}) \cite{yuan2021we}. Intra-client OOD scenarios refer to distribution shifts that occur in unseen tasks within the same client, whereas inter-client OOD scenarios refer to distribution shifts that arise in previously unseen clients.   

Given the long-standing focus on representation learning in machine learning, existing work on OOD generalization in FL primarily concentrates on adopting invariant learning \cite{arjovsky2019invariant,koyama2020invariance,liu2021heterogeneous}, which seeks to learn features that remain consistent across all environments.
In the context of representation learning, the model architecture is typically divided into two distinct components: a feature encoder $\Phi$ to learn representations and a head $\vw$ to get the final predictive outcomes. This can be mathematically represented as $f_\theta=\vw_w \circ \Phi_\phi$, where $\theta=(w,\phi)$. These invariant learning methods operate under the assumption that the representations extracted by the encoder are invariant across all different environments, which can be formalized as:

\begin{assumption}
	\label{invariant}
	There exists a representation $\Phi$ such that for all $e,e' \in \gE_{all}$ and all $\vz$ in the intersection of the supports $Supp(P(\Phi(X^e))) \cap Supp(P(\Phi({X^{e{'}}}))$, we have $$\E[Y^e|\Phi(X^e)=\vz] = \E[{Y^{e{'}}} |\Phi({X^{e{'}}})=\vz].$$
\end{assumption}

Under this assumption, the feature encoder is tasked with managing the heterogeneity among different environments (clients) to learn invariant features. Consequently, the integration of invariant learning within FL frameworks can be uniformly expressed as follows:
\begin{equation}
\label{ood_fl}
    \begin{aligned}
    \centering
        &\min_{\Phi} \sum_{e\in \gE_{train}} \alpha_e R_e(\Phi),
    \end{aligned}
\end{equation}
where $R_e(\Phi)= \frac{1}{|S_e|}\sum_{(x_i)\in S_e}\ell(\Phi(x_i),z)$ denotes the empirical risk of $\Phi$ with invariant features $z$ as labels and $\alpha_e$ denotes the importance weight for environment (client) $e$. \emph{Especially, unlike the empirical risk of the overall model $f$ computing the loss between predicted logits and actual labels $y$, the empirical risk of $\Phi$ calculates using similar or consistent features $z$ (invariant features) as labels, focusing on the feature space.} 
% Based on this framework, various methods have been proposed. 
For instance, some works \cite{guo2023out,tang2023learning} employ the objective (\ref{ood_fl}) using a similar or identical head, while others \cite{zhang2021federated,tan2024heterogeneity} focus on adversarial/contrastive learning to directly optimize the feature encoder. More related work are in Appendix~\ref{relatedwork} 
%Additionally, other studies \cite{deng2020distributionally,zhang2023federated} explore different importance weight strategies to learn more robust features.

% \subsection{OOD Generalization in FedFM}
\subsection{Challenges of OOD Generalization in FedFM}
\label{challenges}
FedFM represents an emerging research area that introduces new challenges beyond those encountered in conventional FL. 
\textbf{(1) Large Parameter Scale:}
Unlike conventional FL focuses on smaller models, like ResNet \cite{he2016deep} with \textasciitilde25 million parameters, FedFM involves foundation models with billions of parameters, such as LLAMA \cite{touvron2023llama} with over 7 billion. This massive scale in FedFM imposes substantial challenges of computation and communication costs during training, making the methods in conventional FL suboptimal for FedFM and necessitating the development of more parameter-efficient learning approaches.
\textbf{(2) Structural Heterogeneity of PEFT Methods:}
Recent research in FedFM adopts PEFT methods \cite{hu2023llm} for efficient learning, freezing most parameters and optimizing only a small subset. While adapting conventional OOD FL methods to PEFT in FedFM can alleviate computation and communication costs, it would face challenges from structural heterogeneity inherent in the varying designs and combinations of PEFT methods \cite{sun2024improving}. For instance, the LoRA method \cite{hu2021lora} in PEFT involves two low-rank matrices that are combined multiplicatively; operating each matrix separately diverges from the objective of jointly optimizing them. Thus, it is essential to develop innovative OOD generalization approaches in FedFM that effectively address the structural heterogeneity of PEFT methods while maintaining efficiency in learning.
\textbf{(3) Increased Data Heterogeneity.}
Foundation models are designed to address a wide range of downstream tasks, leading FedFM to encounter more heterogeneous data than conventional FL \cite{zhuang2023foundation,charles2024towards}. Unlike conventional FL dealing with label or feature distribution heterogeneity, FedFM would encounter cross-dataset or cross-task distribution shifts, collectively referred to as cross-domain distribution heterogeneity. This necessitates personalized models that can effectively adapt to diverse client distributions, thereby enhancing overall performance. However, existing personalization methods in conventional FL often fall short in terms of generalization \cite{jiang2023test,xie2024perada}, making them less effective for versatile applications required in FedFM and highlighting the need for advanced FedFM-specific approaches to achieve better generalization in increased data heterogeneity.

As analyzed above, due to the challenges posed by large parameters, structural heterogeneity and increased data heterogeneity, traditional methods for addressing OOD generalization in conventional FL are inadequate for direct application in FedFM. \emph{This motivates the development of an efficient adapter-based personalized FedFM method with OOD generalization guarantees.}

\section{Method}
To address the above challenges in FedFM, we propose an adapter-based personalized FedFM method with OOD generalization guarantees. In this section, we start by analyzing the generalization bounds of both the conventional global and personalized models in FedFM, then outline our proposed method that facilitates the learning of invariant features through feature distance-based regularization, finally discuss our method's deployment in both intra-client and inter-client OOD scenarios.% Lastly, we discuss the deployment of our method in both intra-client and inter-client OOD scenarios.

\subsection{Generalization Analysis}
\label{sec_gen}
We begin by analyzing the generalization bound of the conventional aggregated global model in FL. The aggregated global hypothesis $f_g$ is defined with the objective $f_g=\argmin_{f\in\gF}\sum_{e\in \gE_{train}}\alpha_e R_e(f)$. Following previous work \cite{konstantinov2019robust}, for any testing environment $e' \in \gE_{all}$, the generalization bound of the global hypothesis $f_g$ is primarily constrained by the discrepancy $\sum_{e\in \gE_{train}}\alpha_ed_\gF(P_e, P_{e{'}})$, where $ d_\gF(P_e, P_{e{'}})=Supp_{f\in \gF}(|\gR_e(f)-\gR_{e{'}}(f)|)$. 

\begin{theorem}
    \label{pos1}
    {\rm(Conventional aggregated global model in FedFM inherently retains OOD generalization ability).}
	In FedFM, we consider learning the global hypothesis $f_g=(\vw,\Phi_g)$. Since foundation models are pre-trained with massive data in one unified format, this results in an optimal and fixed head $\vw$ towards all tasks during tuning \cite{hu2023llm}, that is, $\vw \in \argmin_{\vw} R_e(\vw,\Phi_g)$ for all $e \in \gE_{all}$. Accordingly, the objective of $f_g$ can be further formulated as objective (\ref{ood_fl}) to learn invariant representations $\vz=\Phi_g(X)$. Therefore, the discrepancy $d_\gF(P_e, P_{e{'}})=Supp_{f\in \gF}(|\E[\ell(\vw(\vz)),Y^e]-\E[\ell(\vw(\vz)),Y^{e'}]|)$ approaches zero if $\vz$ is an invariant representation according to Assumption~\ref{invariant}.
\end{theorem}

Due to increased data heterogeneity in FedFM, personalized models are essential to align with the specific distribution of each client for individual user preferences. 
To address this, we further analyze the generalization bound of the conventional personalized model in FedFM. As the head $\vw$ remains fixed during the turning, the difference between personalized hypothesis $f_e=(\vw,\Phi_e)$ and global hypothesis $f_g=(\vw,\Phi_g)$ lies in the feature encoder $\Phi$. 
% Therefore, we can get the upper bound of risk of the personalized model $f_e$ as Theorem~\ref{gen}.

\begin{theorem}
    \label{pos2}
    {\rm(Generalization bound of the personalized model in FedFM is further constrained by the invariant feature distance.)}
	In FedFM, we consider learning the personalized hypothesis $f_e=(\vw,\Phi_e)$. Given that the generalization bound for the global hypothesis $f_g$ has been established in previous work \cite{konstantinov2019robust}, we primarily need to examine the distance $|\gR_{e'}(f_{e})- \gR_{e'}(f_g)|=|\E[\ell(\vw(\Phi_e(X^{e'}))),Y^{e'}]-\E[\ell(\vw(\Phi_g(X^{e'}))),Y^{e'}]|$ to determine the generalization bound for the personalized hypothesis $f_e$. Therefore, based on Assumption~\ref{invariant}, the generalization bound of the personalized model in FedFM is further constrained by $\E[D(\Phi_e(X^{e'}),\Phi_g(X^{e'}))]$, where $D$ denotes the feature distance function.
\end{theorem}

As shown in Theorem~\ref{pos2}, the generalization bound of the conventional personalized model in FedFM is further constrained by the feature distance $\E[D(\Phi_e(X^{e'}),\Phi_g(X^{e'}))]$. Since it is challenging to directly quantify this distance, we are motivated to optimize it during the learning process of the personalized model in FedFM to achieve a tighter generalization bound. 
% However, due to the inaccessibility of unseen environments' data during training, we instead optimize the feature distance using the available training environments and incorporate this distance as a regularization term in the learning of the personalized model. Given that the aggregated global model captures invariant features across all environments, aligning the personalized model's features through this regularization term implicitly encourages the personalized model to align with the global model for invariant feature learning, thereby enhancing its OOD generalization ability. 
For more detailed proofs of the generalization bound, please refer to Appendix \ref{generalization}.

\subsection{Proposed Method}
To enable efficient learning in FedFM, we employ adapter-based PEFT methods \cite{hu2023llm}, where the parameters of foundation models are divided into a majority frozen part and a small tunable part (adapter). During the learning phase in FedFM with PEFT methods, only the adapter is updated and communicated across the federated network to reduce the communication overhead and computational burden.
% As discussed in Section \ref{challenges}, the large parameter scale and increased data heterogeneity present two key challenges for FedFM. 
Additionally, to address the issue of increased data heterogeneity, we introduce an additional personalized adapter for each client, tailored to align with specific data distributions, thereby enhancing overall performance. Simultaneously, to ensure the versatility of foundation models and address the structural heterogeneity of PEFT Methods, we incorporate a feature distance-based regularization term inspired by the generalization analysis in Section~\ref{sec_gen}. This regularization not only leverages insights from the aggregated global model to enhance the OOD generalization of the personalized model, but also implicitly guides the learning of adapter parameters without directly manipulating the adapters themselves to mitigate discordance caused by the diverse structures and combinations in PEFT. 

\paragraph{Optimization Objective.} We focus exclusively on the feature encoder $\Phi$, which consists of tunable adapter $\phi$ and other frozen parts $\phi_{frozen}$, disregarding the fixed head $\vw$. FedOA is designed to learn a personalized $\Phi_e$ for each client, characterized by a unique dataset denoted as $S_e$, while ensuring OOD generalization from the aggregation $\Phi_g$ with regularization,

\begin{equation}
\label{objective}
    \begin{aligned}
    \centering
        &\min_{\Phi_e} & R_e(\Phi_e) + \lambda D(\Phi_e(X^e),\Phi_g^{*}(X^e)) \\
        &s.t.  &\Phi_g^* \in \argmin_{\Phi} \sum_{e\in \gE_{train}} \alpha_e R_e(\Phi_g)
    \end{aligned}
\end{equation}

where $D$ denotes function to measure distance and $\lambda$ controls interpolation between personalized and global models.

% \subsection{Algorithm}
Specifically, as outlined in algorithm~\ref{alg:algorithm}, our method optimizes the personalized and aggregated global adapters iteratively for each round. 
% The following section provides a detailed explanation of our approach. 
\textbf{On the server side}, for each communication round $t \in [T]$, a subset of clients $\gE_t$ is selected. In the first round $t=0$, the server initializes the global adapter $\Phi_g$ with parameters $\phi_g^0$ and broadcasts the initialized global adapter to the selected clients. In subsequent communication rounds $t\in \{1,..,T-1\}$, after receiving the returned global adapter $\phi_g^{t-1,e}$ from each selected client, the server aggregates these adapters across all selected clients to obtain the updated global adapter for the next round, denoted as $\phi_g^{t}=\sum_{e\in \gE_t}\alpha_e\phi_g^{t-1,e}$.
\textbf{On the client side}, each client maintains two adapters: a personalized adapter $\Phi_e$ with parameters $\phi_e$ and a global adapter $\Phi_{g}^e$ with parameters $\phi_{g}^e$. For each communication round $t \in [T]$, the client initializes the personalized adapter as $\phi_{e,0}^{t} = \phi_e^{t-1}$ and performs $K$ local update steps to obtain $\phi_e^t=\phi_{e,K}^{t}$. Similarly, the global adapter in each client is initiated as $\phi_{g}^{e} = \phi_g^{t-1}$ to obtain $\phi_{g}^{t-1,e}$. Especially, the updated global adapter $\phi_{g}^{t-1,e}$ is sent back to the server for aggregation, while the personalized adapter $\phi_e^t$ remains local without communication.

\paragraph{Why feature distance-based regularization?}
\label{sec-regularization}
Compared to conventional FL's parameter regularization methods \cite{li2020federated,li2021ditto,t2020personalized,xie2024perada}, our feature distance-based regularization is better suited for FedFM, effectively addressing the structural heterogeneity of PEFT methods while being more storage- and computation-efficient. First, Unlike parameter regularization, which directly manipulates adapter parameters and risks unintended outcomes (e.g., regularizing each matrix separately of LoRA diverges from the objective of jointly optimizing them), feature distance-based regularization implicitly guides parameter learning, mitigating structural heterogeneity.
Second, feature vectors are much smaller in size compared to the parameters (even adapter parameters) of FedFM, making feature distance-based regularization more storage- and computation-efficient in this context.
% However, for FedFM, parameter regularization would lead to high computation costs and unintended results.Firstly, due to the large scale of parameters in FedFM, applying regularization directly to all parameters incurs substantial computational costs. In contrast, feature vectors are much smaller in size compared to the full parameter set of an FedFM, making feature distance-based regularization more storage- and computation-efficient in this context.Secondly, while parameter regularization could be applied between adapters to reduce computational overhead, it often leads to unintended results due to the varying structures and combinations of PEFT methods used in FedFM as shown in previous work \cite{sun2024improving}. For instance, the LoRA method in PEFT involves two low-rank matrices that are combined multiplicatively; regularizing each matrix separately diverges from the objective of jointly optimizing them. In contrast, feature distance-based regularization avoids this discordance, as it implicitly guides the learning of adapter parameters without directly manipulating the adapters themselves. 
Additionally, unlike previous methods \cite{zhou2023fedfa} that utilize prototypes for regularization requiring a finite categorization, feature distance-based regularization are not bound by a set number of categories and learn invariant features autonomously across different environments by the feature encoder, which is more suitable for federated foundation models in OOD scenarios due to open-vocabulary tasks inherently (e.g. the categories of real-world images are effectively infinite).

\begin{algorithm}[tb]
\caption{FedOA}
\label{alg:algorithm}
\textbf{Input}: Clients $\gE_{train}$, local datasets $\{S_e\}_{e\in \gE_{train}}$, communication rounds $T$, local update steps $K$ \\
% \textbf{Parameter}: Optional list of parameters\\
\textbf{Output}: Personalized adapters $\{\phi_e\}_{e\in \gE_{train}}$ and global adapter $\phi_g$
\begin{algorithmic}[1] %[1] enables line numbers
\FOR{$t=0,...,T-1$}
\STATE Server randomly selects a subset of devices $\gE_t$, and sends $\phi_g^{t-1}$ to them
\FOR{client $e \in \gE_t$ in parallel}
% \FOR{$k=0,...,K-1$}
% \STATE sample mini-batch $\epsilon$ from local data $D_e$
\FOR{$k=0,...,K-1$}
\STATE Sample mini-batch $\xi$ from local data $S_e$
\STATE \textcolor{blue}{// update personalized adapter}
\STATE $\phi_{e,k}^{t}= \phi^{t}_{e,k-1}-\eta_l \nabla( R_e(\phi^{t}_{e,k-1};\xi)+\lambda D(\Phi(\phi^{t}_{e,k-1};\xi), \Phi(\phi^{t-1}_g;\xi)))$
\ENDFOR
\STATE \textcolor{blue}{// update global adapter}
\STATE $\phi_g^{t-1,e}=\phi_g^{t-1}-\eta_g\nabla R_e(\phi_g^{t-1})$ 
\STATE Send $\phi_g^{t-1,e}$ back to server
\ENDFOR
\STATE Server aggregates $\phi_g^{t}=\sum_{e\in \gE_t}\alpha_e\phi_g^{t-1,e}$
\ENDFOR
% \STATE \textbf{return} $\{\phi_e\}_{e\in \gE}$(personalized),$\phi_g$(global)
\end{algorithmic}
\end{algorithm}

\paragraph{Inference.}
As highlighted in previous work \cite{yuan2021we}, OOD scenarios can occur either within the same client (\textbf{intra-client}) or across different clients (\textbf{inter-client}). Intra-client OOD involves test data with distribution shifts from the training data in the same client, while inter-client OOD involves new clients with data distribution differing from training clients. 
% In intra-client OOD scenarios, the test data exhibits distribution shifts from the training data in the same client, while in inter-client OOD scenarios, new clients' data experience distribution shifts from these training clients.
%For example, in a language model deployed for personalized customer support, a client’s training data may consist of conversations focused on troubleshooting technical issues, while the test data might involve inquiries about billing or product returns, introducing a shift in the types of topics and language used. 
% In inter-client OOD scenarios, new clients' data experience distribution shifts from these training clients. 
%For example, in a federated learning system for speech recognition, training clients may consist of users from a specific region with a common accent, but new clients could have distinct accents, leading to a distribution shift in speech patterns.  
Our proposed method could address both: the learned personalized model can be directly deployed to handle the distribution shifts within the same client for intra-client OOD scenarios and the aggregated global model can be deployed to manage distribution shifts among different clients for inter-client OOD scenarios. As analyzed in Section \ref{sec_gen}, conventional aggregation in FedFM is inherently capable of achieving OOD generalization, while conventional personalized adaptation methods often lack this generalization guarantee, resulting in suboptimal performance in intra-client OOD scenarios. Therefore, our experiment primarily focuses on intra-client OOD scenarios to evaluate the effectiveness of the proposed personalized adaptation approach in handling these distribution shifts.

% \section{Theoretical Analysis}

\section{Convergence Analysis}
\label{sec-conv-analysis}
In this section, we delve into the convergence analysis of the proposed method. For the purpose of clarity in our analysis, we restrict our focus to the small tunable part of parameters $\phi$, while excluding other parameters that remain frozen. We first state several standard assumptions on the function.
\begin{assumption}
	\label{a}
    {\rm(Smoothness).}
	For all clients $e$, we assume that $R_e(\phi)$ and $\Phi_e$ are $L$-Lipschitz smoothness, as follows when $\forall \phi,\phi{'}$:
    \begin{equation}
    \begin{aligned}
        ||\nabla R_e(\phi)-\nabla R_e(\phi{'})|| & \leq L||\phi-\phi{'}||, \\
        ||\nabla\Phi_e(\phi)-\nabla\Phi_e(\phi{'})|| & \leq L||\phi-\phi{'}||.
    \end{aligned}
    \end{equation}
\end{assumption}

\begin{assumption}
	\label{b}
	{\rm(Unbiased gradient estimator and Bounded gradients).}
    For all clients $e$, we assume that the expectation of stochastic gradient $\nabla R_e(\phi;\xi)$ and $\nabla\Phi_e(\phi;\xi)$ are unbiased estimators of the local gradients $\nabla R_e(\phi)$ and $\nabla\Phi_e(\phi)$,
    and are uniformly bounded by $\sigma^2$. For $\forall \phi$, we have
    \begin{equation}
    \begin{aligned}
        \E||\nabla R_e(\phi;\xi)||=\nabla R_e(\phi), & \E||\nabla\Phi_e(\phi;\xi)||=\nabla\Phi_e(\phi); \\
        \E||\nabla R_e(\phi;\xi)||^2 \leq \sigma^2, &
        \E||\nabla\Phi_e(\phi;\xi)||^2 \leq \sigma^2.
    \end{aligned}
    \end{equation}
\end{assumption}

\begin{assumption}
	\label{c}
	{\rm(Bounded Diversity).}
    For all clients $e$, we assume that the variance of the local gradient to the global gradient is bounded by $G$. For $\forall e,\phi$, we have
    \begin{equation}
        ||\nabla R_e(\phi)-\nabla R(\phi)||  \leq G.
    \end{equation}
\end{assumption}

Assumption~\ref{a} delineates the smoothness of the local risk function, a technique well-established in the optimization analysis \cite{crane2019dingo,elgabli2022fednew}. Given the dependence of our method on the representation function, we also assume the representation function $\Phi$ is $L$-smoothness. Assumption~\ref{b} establishes a boundary on the variance of the stochastic gradient, an approach commonly used in stochastic optimization analysis \cite{karimireddy2021breaking,wang2021field}. Similarly, we also bound the stochastic gradient of the representation function $\Phi$ in our analysis. 
Assumption~\ref{c} bounds the variance of local gradients relative to the global gradient, a method extensively utilized to quantify statistical heterogeneity in FL \cite{fallah2020personalized}.

For the convenience of analysis, we use L2-distance as the distance function $D$ of the regularization term in equation (\ref{objective}). We now present the convergence results of FedOA for the general non-convex case.

\begin{theorem}
	\label{conv}
	Suppose that Assumption~\ref{a},~\ref{b} and~\ref{c} hold true, our method updates with constant local and global step-size such that $\eta_l \leq \frac{1}{8\sqrt{3(1+3T)T(1+2K)K}\lambda\sigma L}$ and $\eta_g \leq \frac{1}{2\sqrt{6(1+3T)T} L}$. Then, the sequence of iterates generated by our method satisfies:
    \begin{equation}
    \begin{aligned}
     %    \frac{1}{T}\sum_{t=1}^T\E||\nabla\gR_e(\phi_e^{t-1})||^2 \leq & \frac{2(\E\gR_e(\phi_e^{0})-\E\gR_e(\phi_e^{*}))}{T}+\frac{4K(1+2K)(L-1)\sigma^2\eta_l^2}{T} \\
     % & + 96K(1+2K)\lambda^2\sigma^2(L-1)L^2\eta_l^2M^2 \\
     % & + 128K(1+2K)T(1+2T)\lambda^2\sigma^2(L-1)L^2G^2\eta_l^2\eta_g^2
     \frac{1}{T} & \sum_{t=1}^T\E||\nabla R_e(\phi_e^{t-1})||^2  \leq \frac{2(\E R_e(\phi_e^{0})-\E R_e(\phi_e^{*}))}{T} \\
     & + 8K(1+2K)(L-1)(1+12\lambda^2L^2M^2)\sigma^2\eta_l^2 \\
    & + 256K(1+2K)T(1+3T)\lambda^2\sigma^2(L-1)L^2G^2\eta_l^2\eta_g^2.
    \end{aligned}
    \end{equation}
    If we choose the step sizes $\eta_l=\gO(\frac{1}{TKL\sigma})$ and $\eta_g=\gO(\frac{1}{TL})$, we have the convergence rates of our method as
    \begin{equation}
    \begin{aligned}
    \frac{1}{T} &\sum_{t=1}^T\E||\nabla R_e(\phi_e^{t-1})||^2 = \\
    & \gO(\frac{(\E R_e(\phi_e^{0})-\E R_e(\phi_e^{*})}{T}, \frac{1+\lambda^2L^2M^2}{T^2L}, \frac{\lambda^2G^2}{T^2L}).
    \end{aligned}
    \end{equation}
\end{theorem} 

As analyzed above, FedOA converges to a stationary point at a rate of $\gO(\frac{1}{T})$. The heterogeneity between clients and between the personalized and global models is captured by $G$ and $M$, respectively. The impact of these heterogeneities can be reduced by increasing $T$. Similarly, the interpolation between the personalized and global models, controlled by $\lambda$, also becomes less significant as $T$ increases. The full proof of these results is provided in Appendix~\ref{convergence}.

\section{Experiments}
In this section, we present experiments to evaluate the performance of our proposed FedOA method and answer the following questions. \textbf{Q1:} Can the conventional aggregated global model in FedFM demonstrate superior OOD generalization ability compared to the centralized model? \textbf{Q2:} In increased heterogeneity scenarios, can FedOA achieve improved OOD generalization performance relative to existing generalization methods in conventional FL?

\begin{table*}[t]
\caption{OOD results of different models using ``leave-one-task-out'' validation. Centralized and FedIT are tested on a single global model, while the remaining models are tested on personalized models with average results reported. Reading Com represents the Reading Comprehension task.  }
\label{table2}
\begin{center}
\begin{tabular}{l|llll|l}
 \toprule
\bf {Methods} &  {\bf Entailment} & {\bf Sentiment} &  {\bf Paraphrase} & {\bf Reading Com} & {\bf Average} \\
\midrule
Centralized &41.75{\scriptsize$\pm$0.35} & 76.87{\scriptsize$\pm$0.17} & 43.38{\scriptsize$\pm$0.17} & 64.05{\scriptsize$\pm$0.16} & 56.51 \\
FedIT & \textbf{43.00}{\scriptsize$\pm$1.40} & 80.63{\scriptsize$\pm$0.88} & 43.63{\scriptsize$\pm$0.88} & 66.17{\scriptsize$\pm$0.63} & 58.36\\
\midrule
% FedSIM & & & &  & & & & & \\
pFedMe & 37.32{\scriptsize$\pm$1.01} & 75.99{\scriptsize$\pm$0.20} & 44.53{\scriptsize$\pm$0.45} & 50.81{\scriptsize$\pm$0.07} & 52.16 \\
FedLoRA& 41.03{\scriptsize$\pm$1.28} & 78.48{\scriptsize$\pm$0.27} & 43.83{\scriptsize$\pm$0.47} & 64.57{\scriptsize$\pm$1.66} & 56.98\\
PERADA & 36.86{\scriptsize$\pm$0.47}  & 76.45{\scriptsize$\pm$0.69} & 44.24{\scriptsize$\pm$0.10} & 52.36{\scriptsize$\pm$2.29}& 52.48\\
FedSDR &36.70{\scriptsize$\pm$0.49} & 66.90{\scriptsize$\pm$1.05} &43.43{\scriptsize$\pm$0.24} & 41.85{\scriptsize$\pm$1.75}& 47.22\\
\midrule
FedOA & 39.73{\scriptsize$\pm$1.26} & \textbf{82.63}{\scriptsize$\pm$0.59} & \textbf{45.86}{\scriptsize$\pm$0.55} & \textbf{67.96}{\scriptsize$\pm$0.49} & \textbf{59.05}\\
\bottomrule
\end{tabular}
\end{center}
\end{table*}

\subsection{Experiment Setting}
Our framework is flexible and can be adapted to any aggregation algorithm, any adapter-based PEFT method, and any transformer-based foundation model by simply substituting the corresponding components. In this paper, we utilize FedAVG \cite{mcmahan2017communication}, LoRA \cite{hu2021lora}, and large language models (LLMs) \cite{zhao2023survey} as illustrative examples to demonstrate.

\paragraph{Datasets.} We construct four federated datasets, each centered around a distinct task, derived from the Flan \cite{wei2021finetuned}, which encompasses a wide range of NLP tasks from over 60 datasets designed for instruction tuning. The tasks selected include Entailment, Sentiment, Paraphrase and Reading Comprehension, each of which consists of two distinct datasets from different domains, reflecting the increased heterogeneity characteristic of FedFM. 
\emph{Since foundation models standardize all tasks into a uniform format, we can treat all tasks as a single unified task, with the original distinct tasks viewed as different distributions of this unified task.}
Therefore, to better align with OOD settings, we perform the ``leave-one-task-out'' strategy, where one task is set aside as the test environment, while the remaining are used as training environments. ROGUE-1 is used as the evaluation metric and more details are in Appendix~\ref{dataset}.

%\textcolor{red}{What is the setting of federated learning?}

%\subsection{Baselines}
%cite/ other baseline

\paragraph{Baselines and Implementation.} We compare our methods with the following baselines based on the same model architecture: 1) global models: centralized model and FedIT \cite{zhang2023towards}; 2) personalized models: pFedMe \cite{t2020personalized} and FedLoRA \cite{yi2023fedlora}; 3) personalized models with generalization guarantees: PERADA \cite{xie2024perada} and FedSDR \cite{tang2023learning}.  
The centralized model is trained on all data of training environments in one center.
Here, we adapt the training paradigm in pFedMe, FedLoRA, PERADA and FedSDR to federated foundation models with NLP tasks.
%\paragraph{Implementation Details.}
We distribute data between clients based on the dataset for data heterogeneity, with the number of training clients as $|\gE_{train}|=6$. To better evaluate the effectiveness of methods, we assume that all clients are activated for every communication round and set the communication round $T=20$. The alpaca-LoRA\footnote{https://github.com/tloen/alpaca-lora} is adapted as the base model initialized with LLaMA-7B\footnote{https://huggingface.co/huggyllama/llama-7b}. We set $\lambda=0.5$ and choose L2-distance as the distance function $D$. More details about baselines are in Appendix \ref{baselines}.

%For local training, each client conducts 10 local epochs with a batch size of 32. The rank of LoRA is set as r=8r=8 and only applied to \mWq\mW_q and \mWv\mW_v. 
% The updating weight of local LoRA training (FedDPA-T) is $\alpha=0.5$ ($\lambda=0.5$) for federated dataset 1 and $\alpha=0.3$ ($\lambda=0.3$) for federated dataset 2. 
% We set $S=5$ and choose cosine similarity for instance-wise dynamic weighting mechanism. More details are in Appendix~\ref{baselines}.
\begin{table*}[t]
\begin{minipage}[t]{0.48\textwidth}
 \makeatletter\def\@captype{table}
 \caption{\small Ablation study of hyperparameter $\lambda$. RC represents the reading comprehension task.}
\label{table3}
\begin{center}
\begin{tabular}{l|lllll}
 \toprule
\bf {$\lambda$} &  {\bf 0.01} & {\bf 0.1} &  {\bf 0.5} & {\bf 1} & {\bf 2} \\
\midrule
RC & 61.14 & 66.16 & 67.61 & 69.05 & 69.90 \\
\bottomrule
\end{tabular}
\end{center}
\end{minipage}
\hfill
\begin{minipage}[t]{0.48\textwidth}
\makeatletter\def\@captype{table}
\caption{\small Ablation study of different distance function $D$. RC represents reading comprehension task.}
\label{table4}
\begin{center}
\begin{tabular}{l|lll}
 \toprule
\bf {$D$} &  {\bf Cosine} & {\bf Pearson} & {\bf L2} \\
\midrule
RC &51.16 &54.02 & 67.61\\
\bottomrule
\end{tabular}
\end{center}
\end{minipage}
\begin{minipage}{\linewidth}
\begin{minipage}[h]{0.32\linewidth}
\makeatletter\def\@captype{figure}
\centering
% \begin{center}
%\framebox[4.0in]{$\;$}
\includegraphics[width=\linewidth]{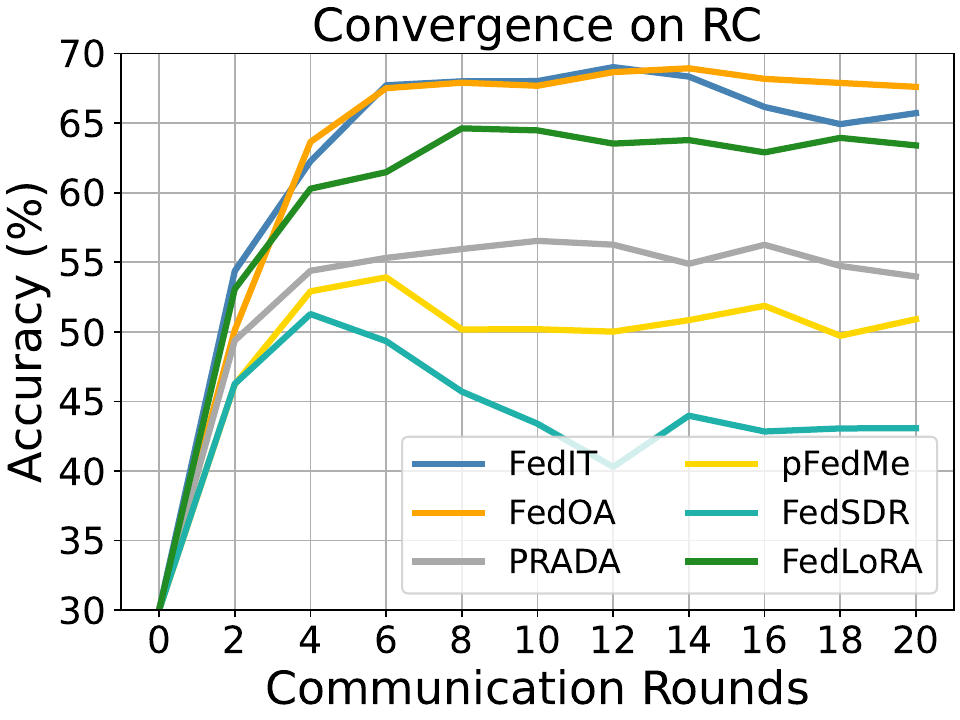} 
% \fbox{\rule[-.5cm]{0cm}{4cm} \rule[-.5cm]{4cm}{0cm}}
\vspace{-20pt}
\caption{\small Average accuracy varies as communication rounds on reading comprehension task.}
\label{fig-conv}
% \end{center}
\end{minipage}
\hfill
\begin{minipage}[h]{0.32\linewidth}
\makeatletter\def\@captype{figure}
\centering
% \begin{center}
%\framebox[4.0in]{$\;$}
\includegraphics[width=\linewidth]{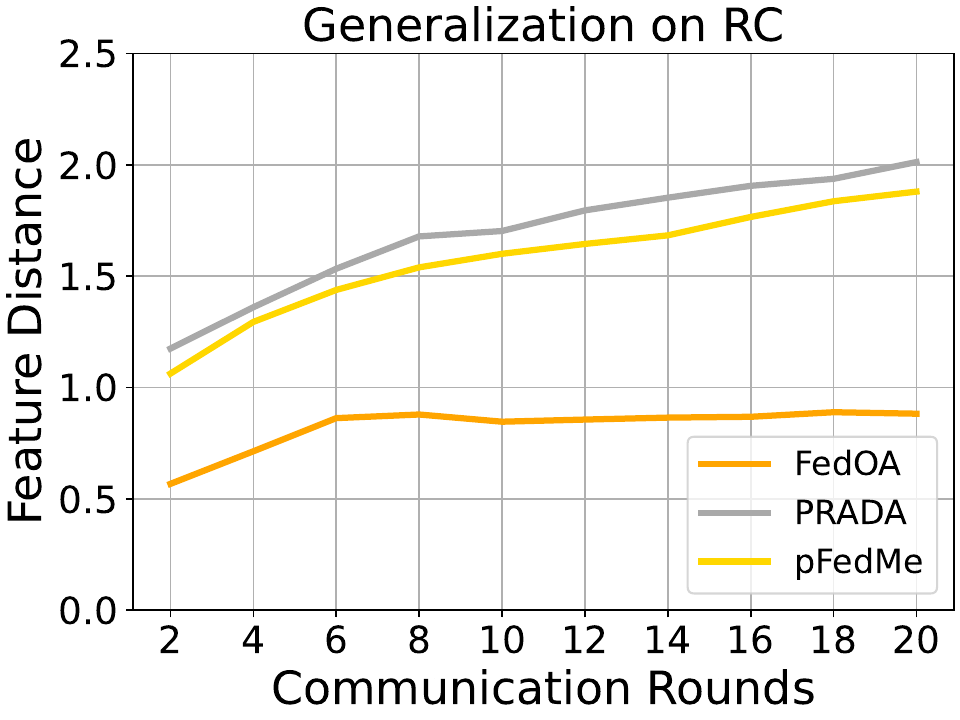} 
% \fbox{\rule[-.5cm]{0cm}{4cm} \rule[-.5cm]{4cm}{0cm}}
\vspace{-20pt}
\caption{\small Feature distance between personalized and global models vs communication rounds.}
% \end{center}
\label{fig-gen-dis}
\end{minipage}
\hfill
\begin{minipage}[h]{0.32\linewidth}
\makeatletter\def\@captype{figure}
\centering
% \begin{center}
%\framebox[4.0in]{$\;$}
\includegraphics[width=\linewidth]{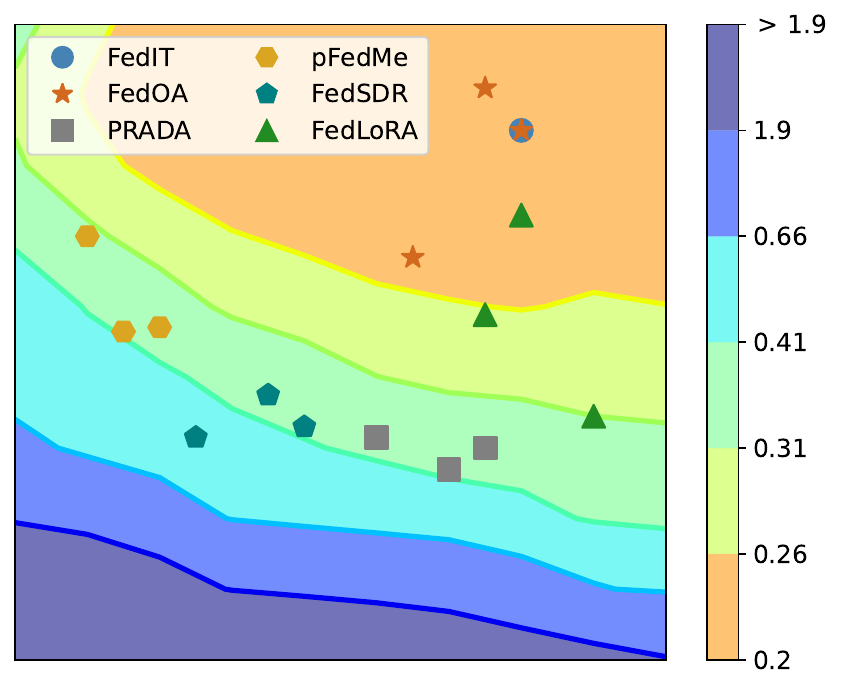} 
% \fbox{\rule[-.5cm]{0cm}{4cm} \rule[-.5cm]{4cm}{0cm}}
\vspace{-20pt}
\caption{\small Loss surfaces w.r.t. model parameters on reading comprehension task.}
% \end{center}
\label{fig-gen-loss}
\end{minipage}
\end{minipage}
\end{table*}
\subsection{Main Results}
\paragraph{Conventional aggregated global model in FedFM achieves better OOD generalization performance than that in centralized setting.} In response to \textbf{Q1}, we compare the OOD generalization performance of the global model in FedFM with that in a centralized setting on four datasets. Specifically, we take FedIT as the baseline method for FedFM to learn the aggregated global model, which adapts FedAVG with LoRA for instruction learning. In this experiment, our proposed FedOA follows the same global model learning process as FedIT, while FedOA is designed to be adaptable to any other global model learning algorithms as well. As shown in Table~\ref{table2}, FedIT exhibits superior OOD generalization performance compared to the centralized model, indicating that conventional aggregation in FedFM can indeed achieve a degree of OOD generalization, consistent with Theorem \ref{pos1}.
% This finding is consistent with the theoretical analysis presented in Theorem \ref{pos1}.  

\paragraph{FedOA demonstrates better OOD generalization performance compared to other baselines.} In response to \textbf{Q2}, we compare FedOA with different baselines on four datasets to assess OOD generalization. Compared to personalized models, as shown in Table~\ref{table2}, FedOA stands out as the most effective among all personalized models, highlighting the importance of feature distance-based regularization from the global adapter for invariant feature learning to improve OOD generalization performance. FedLoRA ranks second, as its further tuning of the learned global model introduces minimal updates, thus maintaining certain OOD generalization ability from the global model. The underperformance of PERADA and pFedMe, which rely on parameter regularization, indicates that this regularization is unsuitable for FedFM due to the discordance between regularization operation and optimization objective. Moreover, the recent benchmark FedSDR for OOD generalization in conventional FL performs poorly, highlighting the inadequacy of conventional FL methods in handling FedFM's increased heterogeneity.
Compared to global models, FedOA leverages the global model's OOD generalization ability to guide personalized models, achieving slightly better average performance than FedIT across four datasets in Table~\ref{table2}. Interestingly, we observe that FedOA outperforms FedIT for most tasks, likely because learning one task would enhance the performance of other tasks with shared underlying knowledge, whereas tasks that vary enormously may lead to degraded performance when learned together \cite{wei2021finetuned}. 

\subsection{Analysis}
\paragraph{Convergence analysis.}
To analyze the convergence of different methods, we examine their average test accuracy versus communication rounds and present the OOD performance comparison on Reading Comprehension in Figure~\ref{fig-conv}. As shown in Figure~\ref{fig-conv}, our method exhibits a convergence speed comparable to other personalized methods, achieving notable performance enhancements after five communication rounds. This aligns with the discussion in Section \ref{sec-conv-analysis}, where FedOA could possess good convergence speed when appropriate learning step sizes are employed. The similar trends observed between our method and FedIT can be attributed to the benefit of feature distance-based regularization from the global adapter for OOD generalization.
% \begin{minipage}{\textwidth}
% \begin{minipage}[t]{0.48\textwidth}
%  \makeatletter\def\@captype{table}
%  \caption{\small Ablation study of hyperparameter $\lambda$. RC represents the reading comprehension task.}
% \label{table3}
% \begin{center}
% \begin{tabular}{l|lllll}
%  \toprule
% \bf {$\lambda$} &  {\bf 0.01} & {\bf 0.1} &  {\bf 0.5} & {\bf 1} & {\bf 2} \\
% \midrule
% RC & 61.14 & 66.16 & 67.61 & 69.05 & 69.90 \\
% \bottomrule
% \end{tabular}
% \end{center}
% \end{minipage}
% \hfill
% \begin{minipage}[t]{0.48\textwidth}
% \makeatletter\def\@captype{table}
% \caption{\small Ablation study of different distance function $D$. RC represents reading comprehension task.}
% \label{table4}
% \begin{center}
% \begin{tabular}{l|lll}
%  \toprule
% \bf {$D$} &  {\bf Cosine} & {\bf Pearson} & {\bf L2} \\
% \midrule
% RC &51.16 &54.02 & 67.61\\
% \bottomrule
% \end{tabular}
% \end{center}
% \end{minipage}
% \end{minipage}

\paragraph{Generalization analysis.}
% \textbf{Loss surface visualization.}
Figure \ref{fig-gen-loss} visualizes the loss surfaces on the test environment for Reading Comprehension, using FedIT's global model as an anchor to position other personalized models. Compared with other methods, FedOA achieves better OOD generalization, as personalized models converge in flatter regions of the loss surface, supporting our theoretical motivation that reducing the distance between global and personalized model features leads to tighter generalization bounds. Additionally, the smaller gaps between global and personalized models highlight FedOA's advantage in maintaining a consistent optimization objective across clients, which is crucial for handling heterogeneous data across diverse domains.
% \textbf{Comparison of invariant feature gaps.}
Figure \ref{fig-gen-dis} compares different regularization terms (feature distance-based regularization of FedOA and parameter regularization of pFedMe and PERADA) based on the average feature distances between personalized models and the global model. FedOA consistently maintains smaller and more stable feature distances, whereas distances in other methods progressively increase, aligning with analysis in Section \ref{sec-regularization} and results in Table \ref{table2}.%demonstrating the effectiveness of our feature distance-based regularization approach. 

% \subsection{Ablation study}

\paragraph{Sensitivity of $\lambda$.}
In this study, we investigated the influence of the hyperparameter $\lambda$ during FedOA training with its value $\lambda \in \{0.01,0.1,0.5,1,2 \}$. As shown in Table \ref{table3}, increasing the regularization weight $\lambda$ will improve the OOD generalization performance, which can be attributed to the greater emphasis on aligning invariant features between the personalized and global models as the regularization strength increases. Notably, even with $\lambda=0.1$, our proposed FedOA achieves superior performance compared to others, which demonstrates the efficiency of our method.

% \begin{table}[t]
% \caption{Ablation study of hyperparameter $\lambda$. RC represents the reading comprehension task.}
% \label{table3}
% \begin{center}
% \begin{tabular}{l|lllll}
%  \toprule
% \bf {$\lambda$} &  {\bf 0.01} & {\bf 0.1} &  {\bf 0.5} & {\bf 1} & {\bf 2} \\
% \midrule
% RC & 61.14 & 66.16 & 67.61 & 69.05 & 69.90 \\
% \bottomrule
% \end{tabular}
% \end{center}
% \end{table}

\paragraph{Effects of different distance function $D$.}
To explore the impact of $D$, we conducted experiments of FedOA with Cosine, Pearson and L2- distance. As shown in Table \ref{table4}, the L2-distance outperforms the others, demonstrating its effectiveness in feature distance calculation. Therefore, we choose the L2-distance function for our feature distance-based regularization during the training of FedOA.

% More experiments and analysis are in Appendix ~\ref{ablations}.
% \begin{table}[t]
% \caption{Ablation study of different distance function $D$. RC represents the reading comprehension task.}
% \label{table4}
% \begin{center}
% \begin{tabular}{l|lll}
%  \toprule
% \bf {$D$} &  {\bf L1} & {\bf L2} & {\bf L$\infty$} \\
% \midrule
% RC & & & \\
% \bottomrule
% \end{tabular}
% \end{center}
% \end{table}

\section{Conclusion}
FedFM offers a promising approach to enhancing foundation models using private data sources, but OOD generalization remains a critical challenge for the FedFM's application across diverse downstream tasks. Previous OOD methods in conventional FL are suboptimal for FedFM due to large parameter scale and increased data heterogeneity. To address these challenges, we begin with a theoretical generalization analysis of FedFM and propose an adapter-based method that incorporates feature distance-based regularization to improve OOD generalization in FedFM, simultaneously providing theoretical convergence guarantees. Our method is evaluated on public NLP tasks simulating an OOD FedFM setting. This work lays the foundation for addressing OOD generalization in FedFM, with future efforts focusing on more advanced methods and larger-scale settings.

\bibliographystyle{plain}
\bibliography{neurips}

%%%%%%%%%%%%%%%%%%%%%%%%%%%%%%%%%%%%%%%%%%%%%%%%%%%%%%%%%%%%

\appendix

\section{Appendix}
The Appendix is organized as follows:
\begin{itemize} 

\item Appendix ~\ref{relatedwork} provides related works.

\item Appendix ~\ref{exp} provides detailed dataset and baseline setups for experiments.

\item Appendix ~\ref{generalization}  provides generalization analysis of FedOA and the full proofs for Theorem~\ref{pos1} and Theorem~\ref{pos2}.

\item Appendix ~\ref{convergence} provides the convergence analysis of FedOA and the full proofs for Theorem~\ref{conv}.

\item Appendix ~\ref{ablations} provides additional experiments demonstrating personalization, scalability and adaptability.

\end{itemize}

\section{Related Work}
\label{relatedwork}
\subsection{Out-of-distribution Generalization}
Out-of-distribution (OOD) generalization addresses scenarios where the distribution of test data differs from that of the training data, a challenge that is critical for the successful deployment of models in real-world applications \cite{liu2021towards,arjovsky2020out}. 
Extensive research has focused on OOD generalization, exploring various assumptions and methodologies. 
% OOD generalization methods can be broadly categorized into three approaches based on differing assumptions and techniques: robust optimization, causal learning, and invariant learning. 
For example, robust optimization methods \cite{namkoong2016stochastic, sagawa2019distributionally,konstantinov2019robust} aim to directly tackle the OOD generalization problem by optimizing for the worst-case error over a set of uncertainty distributions, with constrained relationships between training and testing environments. Causal learning methods \cite{gamella2020active,oberst2021regularizing,yang2021causalvae} draw upon concepts from causal inference to identify and leverage the underlying causal structure of the data, enabling prediction of the outcome variable based on these causal factors. Similarly, invariant learning \cite{arjovsky2019invariant,koyama2020invariance,liu2021heterogeneous} seeks to identify and utilize the underlying heterogeneity and invariant representations or models across different environments by leveraging contextual information.

% However, these methods have primarily been developed within a centralized framework using publicly available data. 
\subsection{Generalization in FL}
Recently, FL has emerged as a promising approach for utilizing private data in model training, prompting increased research into OOD generalization within the FL context \cite{li2023federated,yuan2021we}. 
% In the FL setting, the task in each client can be considered as an environment sampled from a distribution, and OOD generalization aims to train a model that can generalize to unseen environments (new tasks or clients) \cite{yuan2021we}. 
Within this framework, a prevalent approach for achieving OOD generalization in FL is the adaptation of invariant learning based on representation learning. For instance, some studies \cite{zhang2021federated,nguyen2022fedsr,tan2024heterogeneity} employ feature alignment via adversarial/contrastive learning or regularization to align distributions across different clients, facilitating the learning of invariant representations. Similarly, other researchers \cite{guo2023out,tang2023learning} have adapted invariant risk minimization to develop representations that remain invariant to environment-specific variations while retaining relevance for the task at hand. Additionally, given the importance of robust aggregation in FL, numerous studies  \cite{deng2020distributionally,zhang2023federated} have focused on improving aggregation algorithms to enhance OOD generalization. 

Due to the increasing demand for personalized solutions in FL,  recent research has focused on personalized federated learning (PFL) \cite{tan2022towards}, which aims to learn an additional personalized model \cite{t2020personalized,li2021ditto,li2021fedbn} or apply additional personalization steps \cite{fallah2020personalized,collins2021exploiting} to better align with individual user preferences. However, recent studies \cite{jiang2023test,ramasesh2021effect} have revealed that the personalized models in PFL can be prone to catastrophic forgetting and overfitting to local data, thus sacrificing their generalizability. Recent efforts have addressed these challenges by employing techniques such as regularization \cite{zhou2023fedfa,xie2024perada} and designed structure for optimal classifiers \cite{chen2021bridging,luo2022disentangled,li2023no}, but these primarily focus on in-distribution generalization, where only seen training environments are considered during testing. This leaves OOD generalization as a significant unresolved issue in Personalized FL, particularly in the context of FedFM, where models are required to handle various downstream tasks in highly diverse and unseen environments.
% In response to these challenges, the research \cite{xie2024perada} has proposed more efficient PFL frameworks that incorporate parameter-efficient methods to reduce computational and communication costs, while using parameter regularization techniques to preserve generalization ability.
To fill this gap, we investigate the OOD generalization problem within the context of Federated Foundation Models, which are challenged by the substantial computational demands of large parameters and increased data heterogeneity.

\subsection{Federated Foundation Models}
With the advent of foundation models, there has been a growing interest in integrating these models within the FL setting \cite{zhuang2023foundation,yu2023federated,ren2024advances,charles2024towards}. In particular, due to the inherent computational and communication costs, recent research \cite{kuang2023federatedscope,zhang2023fedpetuning} has focused on incorporating adapter-based parameter-efficient tuning (PEFT) methods with federated foundation models. Building on these efforts, numerous studies have emerged to address the challenges of integrating federated foundation models with adapter-based PEFT methods.

One notable contribution \cite{zhang2023towards} pioneered the integration of instruction tuning within federated LLM frameworks. To tackle heterogeneity issues, previous works \cite{babakniya2023slora,cho2023heterogeneous,sun2024improving} introduced novel aggregation and initialization methods for LoRA to enhance the suitability of these models in FL environments. To further optimize the communication and computational overheads of FedFM, other research \cite{xu2023federated,sun2023fedbpt,xu2024fwdllm} has advanced gradient-free optimization techniques that are particularly well-suited for devices with limited memory and computational power.
For personalization, one study \cite{yi2023fedlora} designed a specialized training paradigm for LoRA \cite{hu2021lora} to achieve more effective personalization in visually heterogeneous model scenarios. Additionally, another work \cite{yang2024dual} proposed a dual-adapter framework that incorporates an additional personalized model to enhance personalization efforts. Regarding generalization, a pioneering study \cite{du2024risk} was the first to investigate the generalization degradation that occurs when directly tuning foundation models in FL via robustness analysis experiments.
Diverging from these approaches, our work explores the OOD generalization problem in FedFM through comprehensive theoretical analysis, extending the scope of research in this area.

\section{Implementation Details}
\label{exp}
\subsection{Datasets}
\label{dataset}
In this paper, we developed four datasets derived from the Flan \cite{wei2021finetuned}, and details of their construction are elucidated in this section. Flan comprises a diverse range of NLP tasks, each containing multiple datasets from different domains. To align with OOD settings, we employed a stratified selection process, choosing four distinct tasks to represent four environments and randomly selecting two datasets with different sources for each task from Flan. To simulate client local data scarcity \cite{mcmahan2017communication}, we applied a downsampling strategy, reducing each selected local dataset to 1000 training instances and 200 testing instances. In experiments, we employed a ``leave-one-task-out'' strategy, setting aside one task as the test environment while using the remaining tasks as training environments.
For example, if the task of Entailment (comprising test instances from the snli and anli datasets) is selected as the test dataset, then the remaining six datasets of three tasks (Sentiment, Paraphrase and Reading Comprehension) are used for training with each client contains one dataset.
Consequently, each tested federated OOD dataset encompasses three distinct NLP tasks, with two datasets for each task, yielding a total of 6000 training examples and 1200 testing examples. The specific tasks and datasets included are listed in Table~\ref{table-dataset}. 

\begin{table}[h]
\caption{Tasks and datasets included in the constructed federated OOD datasets.}
\label{table-dataset}
\begin{center}
\begin{tabular}{l|ll}
\toprule
\bf Tasks & \bf Datasets & \bf Sources \\
\midrule
\multirow{2}{*}{Entailment} & snli & Captions\\
 & anli & Wikipedia, WikiHow, news, fiction and formal spoken text\\
 \midrule
\multirow{2}{*}{Sentiment} & sst2 & Movie reviews \\
 & sentiment140 & Tweets\\
 \midrule
 \multirow{2}{*}{Paraphrase} & glue\_mrpc & Newswire articles\\
 & stsb & News headlines, captions and NLI data\\
 \midrule
 Reading  & openbook qa & Wikipedia and ConceptNet\\
 Comprehension& record & CNN/Daily Mail news articles\\
\bottomrule
\end{tabular}
\end{center}
\end{table}

\subsection{Baselines and Implementation}
\label{baselines}
In this section, detailed descriptions of the implementation of each baseline compared in this study will be provided:

\begin{itemize}
\item \textbf{Centralized model:} This model is trained by gathering data from all training environments into a single centralized framework, with 10 epochs to optimize.

\item \textbf{FedIT \cite{zhang2023towards}:} FedIT extends FedAVG \cite{mcmahan2017communication} to foundation models by incorporating LoRA tuning for instruction learning. After training on diverse local client datasets, the final aggregated global model is utilized for testing.

\item \textbf{pFedMe \cite{t2020personalized}:} 
pFedMe learns personalized models through Moreau envelopes regularization. To ensure a fair comparison, we adapt pFedMe to the FedFM setting by incorporating adapter tuning, where only the adapter parameters are learned and regularization is applied specifically to the adapters. 

\item \textbf{FedLoRA \cite{yi2023fedlora}:} FedLoRA incorporates LoRA for efficient learning in model-heterogeneous settings and employs additional local tuning as a personalized adaptation process. Here, we adapt the training paradigm in FedLoRA to NLP tasks, utilizing the personalized LoRAs for testing. These personalized LoRAs are derived through further local tuning on each client's dataset after obtaining the globally aggregated LoRA.

\item \textbf{PERADA \cite{xie2024perada}:}
PERADA utilizes adapters for efficient learning and applies adapter parameter regularization to improve the generalization capability of the personalized model. In this work, we adapt PERADA to the FedFM framework for NLP tasks, excluding the distillation of the global adapter.

\item \textbf{FedSDR \cite{tang2023learning}:} 
FedSDR aims to learn optimal personalized causally invariant predictors through conditional mutual information regularization for addressing OOD scenrios in FL. In this work, we adapt pFedMe to the FedFM setting by incorporating adapter tuning, where only the adapter parameters are learned and regularization is applied specifically to the adapters. Additionally, due to the fixed head in foundation model tuning, we omit the head regularization component typically used for shortcut extractor learning in FedSDR.

\end{itemize}
All models are implemented using LoRA to enhance learning efficiency, with the rank of LoRA set as $r=8$ and only applied to $\mW_q$ and $\mW_v$. For FL methods, each client conducts $K=2$ local epochs with a batch size of 32. We implement all the methods using PyTorch and conduct all experiments on NVIDIA A40 GPUs.

\section{Generalization Analysis}
\label{generalization}

We first analyze the generalization bound of the conventional aggregated global model.
We define the aggregated global hypothesis $f_g$ with its objective as $f_g=\argmin_{f\in\gF}\sum_{e\in \gE_{train}}\alpha_e R_e(f)$. Following previous work \cite{konstantinov2019robust}, we can get the upper bound of risk of the global hypothesis $f_g$ as Lemma~\ref{p1}.

\begin{lemma}
	\label{p1}
    {\rm(Generalization bound of aggregated global).}
	Let $f^*_e=\argmin_{f\in\gF} \gR_e(f)$ and assume that $\ell(.,.)\leq M$, then for any $e \in \gE_{all}$ and $\delta>0$, with probability at least $1-\delta$ over the data, the excess risk of the learned global model $f_g$ can be bounded by:
    \begin{equation}
    \begin{aligned}
        \gR_e(f_g)\leq  \gR_e(f^*_e) +\sum_{e'\in \gE_{train}}\alpha_{e'} H_{e'}(\gF)+2\sum_{e'\in \gE_{train}}\alpha_{e'}d_\gF(P_e, P_{e'})+C\sqrt{\sum_{e'\in \gE_{train}}\frac{\alpha_{e'}}{|S_{e'}|}}
    \end{aligned}
    \end{equation}
    where, $C=6\sqrt{\frac{\log(\frac{4}{\delta})M^2}{2}}$, for each client $e$, $H_e(\gF)$ is the empirical Rademacher complexity $\gF$ and $d_\gF(P_e, P_{e{'}})$ is the discrepancy between the distributions $P_e$ and $P_{e{'}}$ with hypothesis class $\gF$, defined as:
    \begin{equation}
    \begin{aligned}
        d_\gF(P_e, P_{e{'}})=Supp_{f\in \gF}(|\gR_e(f)-\gR_{e{'}}(f)|)
    \end{aligned}
    \end{equation}
\end{lemma}

Following previous work \cite{guo2023out}, we have the definition of invariant predictor (a model only uses invariant features to predict) as Definition~\ref{def1}.

\begin{definition}
    \label{def1}
    {\rm(Invariant Predictor).}
	If there is a head $w$ simultaneously optimal for all environments $\vw \in \argmin_{\vw} R_e(\vw,\Phi)$ for all $e \in \gE_{all}$, the invariant predictor $f=(\vw,\Phi)$ is elicited based on the representation $\Phi$.
\end{definition}

\textbf{Proof of Theorem~\ref{pos1} (Conventional aggregated global model in FedFM inherently retains OOD generalization ability).} During tuning, the pre-trained head $\vw$ of foundation models is fixed and taken as the optimal head for all tasks~\cite{hu2023llm}. Therefore, the objective of global hypothesis $f_g$ can be further formalized as follows: 
\begin{equation}
    \begin{aligned}
    \centering
        &\min_{\Phi_g} & \sum_{e\in \gE_{train}}\alpha_e R_e(\vw,\Phi_g) \\
        &s.t.  & \vw \in \argmin_{\vw} R_e(\vw,\Phi_g), \text{ for all } \textit{} e \in \gE_{train}.
    \end{aligned}
\end{equation}
By omitting the pre-trained head, the objective of global hypothesis $f_g$ simplifies to $\min_{\Phi_g} \sum_{e\in \gE_{train}}\alpha_e R_e(\Phi_g)$, aligning with objective (\ref{ood_fl}) to learn invariant features that satisfy Assumption~\ref{invariant}, according to Definition~\ref{def1}. Hence, based on Lemma~\ref{p1}, when Assumption~\ref{invariant} holds, the discrepancy in the generalization bound of the global hypothesis $f_g$ in federated foundation models approaches zero $d_\gF(P_e, P_{e{'}})=Supp_{f\in \gF}(|\gR_e(f)-\gR_{e{'}}(f)|)=Supp_{f\in \gF}(|\E[\ell(\vw(\vz)),Y^e]-\E[\ell(\vw(\vz)),Y^{e'}]|) \to 0$, and is more tightly bounded by the representation $\Phi$ during learning $d_\gF(P_e, P_{e{'}})=Supp_{f\in \gF}(|\gR_e(f)-\gR_{e{'}}(f)|)=Supp_{\cup\Phi}(|\gR_e(\Phi)-\gR_{e{'}}(\Phi)|)$.

Next, we provide proof of Theorem~\ref{pos2}, where local hypothesis is $f_e=(\vw,\Phi_e)$ and global hypothesis is $f_g=(\vw,\Phi_g)$.
\begin{reptheorem}{pos2}
% \begin{theorem}
	% \label{gen}
    {\rm(Generalization bound of personalized model).}
    Assume that $\ell(.,.)\leq M$ and $f^*_e=\argmin_{f\in\gF}\gR_e(f)$, then for any $e \in \gE_{all}$ and $\delta>0$, with probability at least $1-\delta$ over the data, the excess risk of the learned personalized model $f_e$ can be bounded by:
    \begin{equation}
    \begin{aligned}
        \gR_e(f_e) \leq &  \gR_e(f^*_e) +  M\cdot \E_{X\sim P_e}[D(\Phi_e(X),\Phi_g(X))] +\sum_{e'\in \gE_{train}}\alpha_{e'}H_{e'}(\gF) \\
     &+2\sum_{e'\in \gE_{train}}\alpha_{e'}d_\gF(P_e, P_{e{'}})+C\sqrt{\sum_{e'\in \gE_{train}}\frac{\alpha_{e'}}{|S_{e'}|}}
    \end{aligned}
    \end{equation}
\end{reptheorem}
% \end{theorem}
Proof.
\begin{equation}
    \begin{aligned}
        \gR_e(f_e)=\underbrace{\gR_e(f_e)- \gR_e(f_g)}_{A_1} + \gR_e(f_g)
    \end{aligned}
\end{equation}
Assume $z=\Phi(x)$, for the first term $A_1$, we have
\begin{equation}
    \begin{aligned}
        A_1 =& \E_{z\sim P(\Phi_e(X))}[\E_{y\sim P(Y|Z=z)}[\ell(\vw(z),y)]] - \E_{z{'}\sim P(\Phi_g(X))}[\E_{y\sim P(Y|Z=z{'})}[\ell(\vw(z),y)]] \\
        =& \E_{z\sim P(\Phi_e(X))}[\E_{y\sim P(Y|Z=z)}[\ell(\vw(z),y)]] -\E_{z\sim P(\Phi_e(X))}[\E_{y\sim P(Y|Z=z{'})}[\ell(\vw(z),y)]] \\
        &+ \E_{z\sim P(\Phi_e(X))}[\underbrace{\E_{y\sim P(Y|Z=z{'})}[\ell(\vw(z),y)]}_{g(z)}] - \E_{z{'}\sim P(\Phi_g(X))}[\underbrace{\E_{y\sim P(Y|Z=z{'})}[\ell(\vw(z),y)]}_{g(z)}] \\
        \overset{(a)}{\leq} &  \E_{z\sim P(\Phi_e(X))}[g(z)]-\E_{z{'}\sim P(\Phi_g(X))}[g(z)] \\
        \overset{(b)}{\leq} & M\cdot \E_{X\sim P_e}[D(\Phi_e(X),\Phi_g(X))]
    \end{aligned}
\end{equation}
where (a) is from Assumption~\ref{invariant}, (b) is from the condition that $|g(z)|\leq M$ if $ \ell(.,.)\leq M$, and $D$ represents a function to measure distance. 

Plugging back the bounds on $A_1$ and Lemma~\ref{p1}, obtaining
\begin{equation}
    \begin{aligned}
     \gR_e(f_e) \leq &  \gR_e(f^*_e) +  M\cdot \E_{X\sim P_e}[D(\Phi_e(X),\Phi_g(X))] +\sum_{e'\in \gE_{train}}\alpha_{e'}H_{e'}(\gF) \\
     &+2\sum_{e'\in \gE_{train}}\alpha_{e'}d_\gF(P_e, P_{e{'}})+C\sqrt{\sum_{e'\in \gE_{train}}\frac{\alpha_{e'}}{|S_{e'}|}}
    \end{aligned}
\end{equation}

\section{Convergence Analysis}
\label{convergence}
\subsection{Technical Lemmas}
We first present some technical lemmas involved in later proofs, where Lemma~\ref{l1} and Lemma~\ref{l2} can be found in \cite{karimireddy2020scaffold} and \cite{t2020personalized}, respectively.

\begin{lemma}
	\label{l1}
	{\rm(Relaxed triangle inequality).}
    For any vectors $v_1,v_2 \in \sR^d$ and $a >0 $, we have
    \begin{equation}
        ||v_1+v_2||^2 \leq (1+a)||v_1||^2+(1+\frac{1}{a})||v_2||^2.
    \end{equation}
\end{lemma}

\begin{lemma}
	\label{l2}
	{\rm(Relaxed triangle inequality).}
    For any $x \in \sR, n \in \sN $, we have
    \begin{equation}
    \begin{aligned}
        \sum_{i=0}^{N-1}x^i & =\frac{x^n-1}{x-1}, \\
        (1+\frac{x}{n})^n & \leq e^x
    \end{aligned}
    \end{equation}
\end{lemma}

\begin{lemma}
	\label{l-div}
	{\rm(Heterogenity Bound).}
    Suppose that Assumption~\ref{c} holds true, we have
    \begin{equation}
    \begin{aligned}
        \E||\nabla  R(\phi)||^2 \leq 2\E||\nabla R_e(\phi)||^2 + 2G^2 
    \end{aligned}
    \end{equation}
\end{lemma}
Proof. Using Lemma~\ref{l1} and Assumption~\ref{c}, we have
    \begin{equation}
    \begin{aligned}
        \E||\nabla  R(\phi)||^2 =& \E||\nabla  R(\phi)-\nabla  R_e(\phi) + \nabla  R_e(\phi)||^2 \\
        \leq &2\E||\nabla R_e(\phi)||^2 + 2G^2 
    \end{aligned}
    \end{equation}

% \begin{lemma}
% 	\label{pl}
% 	{\rm($\mu$-PL Condition).}
%     A function is $\mu$-PL if
%     \begin{equation}
%     \begin{aligned}
%         ||\nabla f(x)||^2 \geq \mu(f(x)-f^*)
%     \end{aligned}
%     \end{equation}
% \end{lemma}
\subsection{Convergence Results}
In this section, we provide proof of Theorem~\ref{conv}, focusing exclusively on the small tunable parameter $\phi$, while disregarding the frozen parameters.

We begin by defining the local updates for each client $e$. The client's global model, with parameter $\phi^{t-1}_g$, and the personalized model, initialized with $\phi^{t}_{e,0}=\phi^{t-1}_e$, undergo K local updates with L2-distance function $D$, as follows:
\begin{equation}
\begin{aligned}
    \phi^{t}_{e,k} & = \phi^{t}_{e,k-1}-\eta_l g_e(\phi^{t}_{e,k-1}, \phi^{t-1}_g) \\ 
    & = \phi^{t}_{e,k-1}-\eta_l [\nabla R_e(\phi^{t}_{e,k-1};\xi)+\lambda\nabla D(\Phi(\phi^{t}_{e,k-1};\xi), \Phi(\phi^{t-1}_g;\xi))] \\
    & = \phi^{t}_{e,k-1} -\eta_l [\nabla R_e(\phi^{t}_{e,k-1};\xi)+2\lambda\nabla\Phi(\phi^{t}_{e,k-1};\xi)|\Phi(\phi^{t}_{e,k-1};\xi)-\Phi(\phi^{t-1}_g;\xi)| ]
\end{aligned}
\end{equation}

We then bound the client drift error.
\begin{lemma}
	\label{l3}
	Suppose that Assumption~\ref{a} and ~\ref{b} hold true, our method updates with constant local step-size such that $\eta_l \leq \frac{1}{4\sqrt{2(1+2K)K}\lambda\sigma L}$. Then, for all $t \in [T]$, we can bound the client drift error as follows:
    \begin{equation}
    \begin{aligned}
        \E||\phi^t_{e,K}-\phi^t_{e,0}||^2 \leq 32K(1+2K)\lambda^2\sigma^2L^2\eta_l^2\E||\phi^{t}_{e,0}-\phi^{t-1}_g||^2 + 4K(1+2K)\sigma^2\eta_l^2
    \end{aligned}
    \end{equation}
\end{lemma}
Proof. 
\begin{equation}
\begin{aligned}
\E||\phi^t_{e,K}-\phi^t_{e,0}||^2 & = \E||\phi^t_{e,K-1}-\phi^t_{e,0}-\eta_lg_c(\phi^t_{e,K-1},\phi^{t-1}_g)||^2 \\
& \overset{(a)}{\leq} (1+\frac{1}{2K})\E||\phi^t_{e,K-1}-\phi^t_{e,0}||^2 + \underbrace{(1+2K)\eta_l^2\E||g_c(\phi^t_{e,K-1},\phi^{t-1}_g)||^2}_{A_1}
\end{aligned}
\end{equation}
where (a) is from Lemma~\ref{l1} with $a=2K$. For the second term, we have
\begin{equation}
\begin{aligned}
A_1 = & (1+2K)\eta_l^2\E||\nabla R_e(\phi^{t}_{e,K-1};\xi)+2\lambda\nabla\Phi(\phi^{t}_{e,K-1};\xi)||\Phi(\phi^{t}_{e,K-1};\xi)-\Phi(\phi^{t-1}_g;\xi)||||^2 \\
\overset{(b)}{\leq} & 2(1+2K)\eta_l^2\E||\nabla R_e(\phi^{t}_{e,K-1};\xi)||^2 \\
& +8(1+2K)\lambda^2\eta_l^2\E[||\nabla\Phi(\phi^{t}_{e,K-1};\xi)||^2\cdot||\Phi(\phi^{t}_{e,K-1};\xi)-\Phi(\phi^{t-1}_g;\xi)||^2] \\
\overset{(c)}{\leq} & 2(1+2K)\sigma^2\eta_l^2+8(1+2K)\lambda^2\sigma^2L^2\eta_l^2\E||\phi^{t}_{e,K-1}-\phi^{t-1}_g||^2 \\
\overset{(d)}{\leq} &
2(1+2K)\sigma^2\eta_l^2+16(1+2K)\lambda^2\sigma^2L^2\eta_l^2\E||\phi^{t}_{e,K-1}-\phi^{t}_{e,0}||^2 \\
& +16(1+2K)\lambda^2\sigma^2L^2\eta_l^2\E||\phi^{t}_{e,0}-\phi^{t-1}_g||^2
\end{aligned}
\end{equation}
where (b) is from Lemma~\ref{l1} with $a=1$, (c) is from Assumption~\ref{a} and Assumption~\ref{b}, and (d) is from Lemma~\ref{l1} with $a=1$. Plugging back the bounds on $A_1$, we obtain the recursive bound of the client drift error:
\begin{equation}
\begin{aligned}
\E||\phi^t_{e,K}-\phi^t_{e,0}||^2 \leq & (1+\frac{1}{2K} + 16(1+2K)\lambda^2\sigma^2L^2\eta_l^2)\E||\phi^t_{e,K-1}-\phi^t_{e,0}||^2 \\
& + 16(1+2K)\lambda^2\sigma^2L^2\eta_l^2\E||\phi^{t}_{e,0}-\phi^{t-1}_g||^2 + 2(1+2K)\sigma^2\eta_l^2 \\
\overset{(e)}{\leq} &
(1+\frac{1}{K})\E||\phi^t_{e,K-1}-\phi^t_{e,0}||^2 + 16(1+2K)\lambda^2\sigma^2L^2\eta_l^2\E||\phi^{t}_{e,0}-\phi^{t-1}_g||^2 \\
& + 2(1+2K)\sigma^2\eta_l^2 \\
\overset{(f)}{\leq} &
(16(1+2K)\lambda^2\sigma^2L^2\eta_l^2\E||\phi^{t}_{e,0}-\phi^{t-1}_g||^2 + 2(1+2K)\sigma^2\eta_l^2)\sum_{i=0}^{K-1}(1+\frac{1}{K})^i \\
\overset{(g)}{\leq} &
32K(1+2K)\lambda^2\sigma^2L^2\eta_l^2\E||\phi^{t}_{e,0}-\phi^{t-1}_g||^2 + 4K(1+2K)\sigma^2\eta_l^2
\end{aligned}
\end{equation}
where (e) is from the condition on local step-size that $\eta_l \leq \frac{1}{4\sqrt{2(1+2K)K}\lambda\sigma L}$ implying that $16(1+2K)\lambda^2\sigma^2L^2\eta_l^2\leq \frac{1}{2K}$, (f) is from the unrolling recursion, and (g) is from Lemma~\ref{l2} with $\sum_{i=0}^{K-1}(1+\frac{1}{K})^i=\frac{(1+1/K)^K-1}{1/K}\leq \frac{e-1}{1/K}\leq 2K$.

\begin{lemma}
	\label{l5}
	Suppose that Assumption~\ref{a},~\ref{b} and~\ref{c} hold true, our method updates with constant local and global step-size such that $\eta_l \leq \frac{1}{8\sqrt{3(1+3T)T(1+2K)K}\lambda\sigma L}$ and $\eta_g \leq \frac{1}{2\sqrt{6(1+3T)T} L}$. Then, we have:
    \begin{equation}
    \begin{aligned}
        \E||\phi^{t}_{e}-\phi^{t}_g||^2 \leq 3\E||\phi^{0}_{e}-\phi^{0}_g||^2 + 16(1+3T)TK(1+2K)\sigma^2\eta_l^2 +8(1+3T)T\eta_g^2G^2
    \end{aligned}
    \end{equation}
\end{lemma}
Proof.
\begin{equation}
    \begin{aligned}
    \E||\phi^{t}_{e}-\phi^{t}_g||^2 = & \E||\phi^{t-1}_{e}-\phi^{t-1}_g+\phi^{t}_{e}-\phi^{t-1}_{e}+\phi^{t-1}_g-\phi^{t}_g||^2 \\
\overset{(a)}{\leq} & (1+\frac{1}{3T})\E||\phi^{t-1}_{e}-\phi^{t-1}_g||^2  +\underbrace{(1+3T)\E||\phi^{t}_{e}-\phi^{t-1}_{e}+\phi^{t-1}_g-\phi^{t}_g||^2}_{A_1} \\
    \end{aligned}
\end{equation}
where (a) is from Lemma~\ref{l1} with $a=3T$. For the second term, we have
\begin{equation}
    \begin{aligned}
    A_1 =  & (1+3T)\E||\phi^{t}_{e}-\phi^{t-1}_{e}+\phi^{t-1}_g-\phi^{t}_g||^2 \\
\overset{(b)}{\leq} & 2(1+3T)\E||\phi^{t}_{e}-\phi^{t-1}_{e}||^2 +2(1+3T)\eta_g^2\E||\nabla R(\phi_g^{t-1})||^2 \\
\overset{(c)}{\leq} & 2(1+3T)\E||\phi^{t}_{e}-\phi^{t-1}_{e}||^2 + 4(1+3T)\eta_g^2\E||\nabla R_e(\phi_g^{t-1})||^2+4(1+3T)\eta_g^2G^2\\
\overset{(d)}{\leq} & 2(1+3T)\E||\phi^{t}_{e}-\phi^{t-1}_{e}||^2 + 8(1+3T)\eta_g^2\E||\nabla R_e(\phi_g^{t-1})-\nabla R_e(\phi_e^{t-1})||^2 \\
&+ 8(1+3T)\eta_g^2\E||\nabla R_e(\phi_e^{t-1})||^2 +4(1+3T)\eta_g^2G^2 \\
\overset{(e)}{\leq} & 64(1+3T)K(1+2K)\lambda^2\sigma^2L^2\eta_l^2\E||\phi^{t-1}_{e}-\phi^{t-1}_g||^2 + 8(1+3T)K(1+2K)\sigma^2\eta_l^2 \\
&+ 8(1+3T)L^2\eta_g^2\E||\phi_e^{t-1}-\phi_g^{t-1}||^2 + 8(1+3T)\sigma^2\eta_g^2 +4(1+3T)\eta_g^2G^2\\
    \end{aligned}
\end{equation}
where (b) is from Lemma~\ref{l1} with $a=1$, (c) is from Lemma~\ref{l-div}, (d) is from Lemma~\ref{l1} with $a=1$, (e) is from Lemma~\ref{l3} with $\phi_e^{t-1}=\phi_{e,0}^{t}, \phi_e^{t}=\phi_{e,K}^t$ and Assumption~\ref{a} and Assumpation~\ref{b}. Plugging back the bounds on $A_1$, we obtain the recursive bound as:
\begin{equation}
\begin{aligned}
 \E||\phi^{t}_{e}-\phi^{t}_g||^2 \leq & (1+\frac{1}{3T})\E||\phi^{t-1}_{e}-\phi^{t-1}_g||^2 + 64(1+3T)K(1+2K)\lambda^2\sigma^2L^2\eta_l^2\E||\phi^{t-1}_{e}-\phi^{t-1}_g||^2 \\
& + 8(1+3T)K(1+2K)\sigma^2\eta_l^2 + 8(1+3T)L^2\eta_g^2\E||\phi_e^{t-1}-\phi_g^{t-1}||^2 \\
&+ 8(1+3T)\sigma^2\eta_g^2 +4(1+3T)\eta_g^2G^2\\
\overset{(f)}{\leq} & (1+\frac{1}{T})\E||\phi^{t-1}_{e}-\phi^{t-1}_g||^2 + 8(1+3T)K(1+2K)\sigma^2\eta_l^2 +4(1+3T)\eta_g^2G^2  \\
\overset{(g)}{\leq} & (8(1+3T)K(1+2K)\sigma^2\eta_l^2 +4(1+3T)\eta_g^2G^2)\sum_{i=0}^{T-1}(1+\frac{1}{T})^i + (1+\frac{1}{T})^T\E||\phi^{0}_{e}-\phi^{0}_g||^2 \\
\overset{(h)}{\leq} & 3\E||\phi^{0}_{e}-\phi^{0}_g||^2 + 16(1+3T)TK(1+2K)\sigma^2\eta_l^2 +8(1+3T)T\eta_g^2G^2
\end{aligned}
\end{equation}
where (f) is from the condition on global step-size that $\eta_g \leq \frac{1}{2\sqrt{6(1+3T)T} L}$ implying that $8(1+3T)L^2\eta_g^2\leq \frac{1}{3T}$, and local step-size that $\eta_l \leq \frac{1}{8\sqrt{3(1+3T)T(1+2K)K}\lambda\sigma L}$ implying that $64(1+3T)K(1+2K)\lambda^2\sigma^2L^2\eta_l^2\leq \frac{1}{3T}$, (g) is from the unrolling recursion, and (h) is from Lemma~\ref{l2}.

Next, we prove the progress made in each round.
\begin{lemma}
	\label{l4}
	Suppose that Assumption~\ref{a},~\ref{b} and~\ref{c} hold true, our method updates with constant local and global step-size such that $\eta_l \leq \frac{1}{8\sqrt{3(1+3T)T(1+2K)K}\lambda\sigma L}$ and $\eta_g \leq \frac{1}{2\sqrt{6(1+3T)T} L}$. Then, our method makes progress in each round as follows:
    \begin{equation}
    \begin{aligned}
        % \E R_e(\phi^t_e) \leq & \E R_e(\phi_e^{t-1})- \frac{1}{2}||\nabla R_e(\phi_e^{t-1})||^2+ 16K(1+2K)\lambda^2\sigma^2(L-1)L^2\eta_l^2\E||\phi^{t-1}_{e}-\phi^{t-1}_g||^2 \\ & + 2K(1+2K)(L-1)\sigma^2\eta_l^2
        \E R_e(\phi^t_e) \leq & \E R_e(\phi_e^{t-1})- \frac{1}{2}||\nabla R_e(\phi_e^{t-1})||^2+ 48K(1+2K)\lambda^2\sigma^2(L-1)L^2\eta_l^2M^2 \\
& + 128K(1+2K)T(1+3T)\lambda^2\sigma^2(L-1)L^2G^2\eta_l^2\eta_g^2 + 4K(1+2K)(L-1)\sigma^2\eta_l^2
    \end{aligned}
    \end{equation}
\end{lemma}
Proof. Starting from the smoothness, we have
\begin{equation}
\begin{aligned}
\E R_e(\phi^t_e) \leq & \E R_e(\phi_e^{t-1})+\E \langle \nabla R_e(\phi_e^{t-1}), \phi^t_e-\phi_e^{t-1} \rangle+\frac{L}{2}\E||\phi^t_e-\phi_e^{t-1}||^2 \\
\overset{(a)}{\leq} & \E R_e(\phi_e^{t-1})+\frac{L}{2}\E||\phi^t_e-\phi_e^{t-1}||^2 - \frac{1}{2}||\nabla R_e(\phi_e^{t-1})||^2 - \frac{1}{2}\E||\phi^t_e-\phi_e^{t-1}||^2 \\
%\overset{(b)}{\leq} & \E R_e(\phi_e^{t-1})- \frac{1}{2}||\nabla R_e(\phi_e^{t-1})||^2+\frac{L-1}{2}(32K(1+2K)\lambda^2\sigma^2L^2\eta_l^2\E||\phi^{t}_{e,0}-\phi^{t-1}||^2 + 4K(1+2K)\sigma^2\eta_l^2) \\
\overset{(b)}{\leq} & \E R_e(\phi_e^{t-1})- \frac{1}{2}||\nabla R_e(\phi_e^{t-1})||^2+ 16K(1+2K)\lambda^2\sigma^2(L-1)L^2\eta_l^2\E||\phi^{t-1}_{e}-\phi^{t-1}_g||^2 \\
& + 2K(1+2K)(L-1)\sigma^2\eta_l^2 \\
\overset{(c)}{\leq} & \E R_e(\phi_e^{t-1})- \frac{1}{2}||\nabla R_e(\phi_e^{t-1})||^2+ 48K(1+2K)\lambda^2\sigma^2(L-1)L^2\eta_l^2M^2 \\
& + 128K(1+2K)T(1+3T)\lambda^2\sigma^2(L-1)L^2G^2\eta_l^2\eta_g^2 + 4K(1+2K)(L-1)\sigma^2\eta_l^2
\end{aligned}
\end{equation}
where (a) is from that $-\langle \va,\vb \rangle \leq \frac{1}{2}(||\va||^2+||\vb||^2)$, (b) is from that $\phi^{t}_{e,0}=\phi^{t-1}_e$ and substituting with Lemma~\ref{l3}, and (c) is from that $\E||\phi^{0}_{e}-\phi^{0}||^2 \leq M^2$ and substituting with Lemma~\ref{l5}

% Convergence results for $\mu$-PL inequality case
% \begin{theorem}
% 	\label{l6}
% 	Suppose that Assumption~\ref{a},~\ref{b} and~\ref{c} hold true, our method updates with constant local and global step-size such that $\eta_l \leq \frac{1}{4\sqrt{2(1+2K)K}\lambda\sigma L}$ and $\eta_g \leq \frac{1}{4\sqrt{(1+2T)T}\lambda\sigma L}$. Then, the sequence of iterates generated by our method satisfies:
%     \begin{equation}
%     \begin{aligned}
%      \E[ R(\phi_e)- R^*] \leq
%     \end{aligned}
%     \end{equation}
% \end{theorem}
% Proof. Using the $\mu$-PL inequality to, we have

Finally, we can get convergence results for the general non-convex case of our method.
\begin{reptheorem}{conv}
	% \label{l6}
	Suppose that Assumption~\ref{a},~\ref{b} and~\ref{c} hold true, our method updates with constant local and global step-size such that $\eta_l \leq \frac{1}{8\sqrt{3(1+3T)T(1+2K)K}\lambda\sigma L}$ and $\eta_g \leq \frac{1}{2\sqrt{6(1+3T)T} L}$. Then, the sequence of iterates generated by our method satisfies:
    \begin{equation}
    \begin{aligned}
     %    \frac{1}{T}\sum_{t=1}^T\E||\nabla R_e(\phi_e^{t-1})||^2 \leq & \frac{2(\E R_e(\phi_e^{0})-\E R_e(\phi_e^{*}))}{T}+\frac{4K(1+2K)(L-1)\sigma^2\eta_l^2}{T} \\
     % & + 96K(1+2K)\lambda^2\sigma^2(L-1)L^2\eta_l^2M^2 \\
     % & + 128K(1+2K)T(1+2T)\lambda^2\sigma^2(L-1)L^2G^2\eta_l^2\eta_g^2
     \frac{1}{T}\sum_{t=1}^T\E||\nabla R_e(\phi_e^{t-1})||^2 \leq & \frac{2(\E R_e(\phi_e^{0})-\E R_e(\phi_e^{*}))}{T} + 8K(1+2K)(L-1)(1+12\lambda^2L^2M^2)\sigma^2\eta_l^2 \\
    & + 256K(1+2K)T(1+3T)\lambda^2\sigma^2(L-1)L^2G^2\eta_l^2\eta_g^2
    \end{aligned}
    \end{equation}
    If we choose the step sizes $\eta_l=\gO(\frac{1}{TKL\sigma})$ and $\eta_g=\gO(\frac{1}{TL})$, we have the convergence rates of our method as follows
    \begin{equation}
    \begin{aligned}
    \frac{1}{T}\sum_{t=1}^T\E||\nabla R_e(\phi_e^{t-1})||^2 = \gO(\frac{(\E R_e(\phi_e^{0})-\E R_e(\phi_e^{*})}{T}, \frac{1+\lambda^2L^2M^2}{T^2L}, \frac{\lambda^2G^2}{T^2L})
    \end{aligned}
    \end{equation}
\end{reptheorem}
Proof. Summing up all the $T$ inequalities in Equation of Lemma~\ref{l4}, we have
% \begin{equation}
% \begin{aligned}
%      \frac{1}{T}\sum_{t=1}^T\E||\nabla R_e(\phi_e^{t-1})||^2 \leq & \frac{2\sum_{t=1}^T(\E R_e(\phi_e^{t-1})-\E R_e(\phi_e^{t}))}{T} \\
%      & + 32K(1+2K)\lambda^2\sigma^2(L-1)L^2\eta_l^2\frac{\sum_{t=1}^T\E||\phi^{t-1}_{e}-\phi^{t-1}||^2}{T} \\
%      & +\frac{4K(1+2K)(L-1)\sigma^2\eta_l^2}{T} \\
%      \overset{(a)}{\leq} & \frac{2(\E R_e(\phi_e^{0})-\E R_e(\phi_e^{*}))}{T} \\
%      & + 32K(1+2K)\lambda^2\sigma^2(L-1)L^2\eta_l^2\frac{\sum_{t=1}^T\E||\phi^{t-1}_{e}-\phi^{t-1}||^2}{T} \\
%      & +\frac{4K(1+2K)(L-1)\sigma^2\eta_l^2}{T} \\
%      \overset{(b)}{\leq} &  \frac{2(\E R_e(\phi_e^{0})-\E R_e(\phi_e^{*}))}{T}+\frac{4K(1+2K)(L-1)\sigma^2\eta_l^2}{T} \\
%      & + 96K(1+2K)\lambda^2\sigma^2(L-1)L^2\eta_l^2M^2 \\
%      & + 128K(1+2K)T(1+2T)\lambda^2\sigma^2(L-1)L^2G^2\eta_l^2\eta_g^2
% \end{aligned}
% \end{equation}
\begin{equation}
    \begin{aligned}
    \frac{1}{T}\sum_{t=1}^T\E||\nabla R_e(\phi_e^{t-1})||^2 \leq & \frac{2\sum_{t=1}^T(\E R_e(\phi_e^{t-1})-\E R_e(\phi_e^{t}))}{T} 
    + 8K(1+2K)(L-1)(1+12\lambda^2L^2M^2)\sigma^2\eta_l^2 \\
    & + 256K(1+2K)T(1+3T)\lambda^2\sigma^2(L-1)L^2G^2\eta_l^2\eta_g^2 \\
    \overset{(a)}{\leq} & \frac{2(\E R_e(\phi_e^{0})-\E R_e(\phi_e^{*}))}{T} + 8K(1+2K)(L-1)(1+12\lambda^2L^2M^2)\sigma^2\eta_l^2 \\
    & + 256K(1+2K)T(1+3T)\lambda^2\sigma^2(L-1)L^2G^2\eta_l^2\eta_g^2
    \end{aligned}
\end{equation}
where (a) is from that $\E R_e(\phi_e^{*})\leq \E R_e(\phi_e^{T})$.

\section{Additional Experiments}
\label{ablations}
\subsection{Personalization Analysis}
As our method considers the OOD generalization of personalized models, we further analyze its personalization performance. As shown in Table~\ref{table-personalization}, personalized methods generally outperform the global aggregated model FedIT, with our proposed method achieving the second-best performance, only marginally lower (by 0.33\%) than the top-performing method, PERADA. These results demonstrate that our approach achieves superior OOD generalization without compromising personalization performance, striking a balance between these two critical objectives.

\begin{table}[h]
\caption{Ablation study of personalization experiments across three tasks. FedIT is tested on a single global model, while the remaining models are tested on personalized models with average results reported.}
\label{table-personalization}
\begin{center}
\begin{tabular}{l|lll|l}
 \toprule
\bf {Methods} &  {\bf Entailment} & {\bf Sentiment} &  {\bf Paraphrase} & {\bf Average} \\
\midrule
FedIT & 64.75 & 82.75 & 59.75 & 69.08 \\
\midrule
% FedSIM & & & &  & & & & & \\
pFedMe & 67.15 & 83.25 & 62.25 & 70.88 \\
FedLoRA& 66.50 & 82.50 & 62.75 & 70.58 \\
PERADA & 69.25 & 83.00 & 62.50 & \textbf{71.58} \\
FedSDR & 66.25 & 82.00 & 62.75 & 70.33 \\
\midrule
FedOA & 69.50 & 82.25 & 62.00 & \underline{71.25} \\
\bottomrule
\end{tabular}
\end{center}
\end{table}

% \begin{table}[h]
% \caption{Ablation study of scalability with 30 clients using ``leave-one-task-out'' validation. FedIT are tested on a single global model, while the remaining models are tested on personalized models with average results reported. Reading Com represents the Reading Comprehension task.  }
% \label{table-p}
% \begin{center}
% \begin{tabular}{l|llll|l}
%  \toprule
% \bf {Methods} &  {\bf Entailment} & {\bf Sentiment} &  {\bf Paraphrase} & {\bf Reading Com} & {\bf Average} \\
% \midrule
% FedIT & 77.00 & 72.92 & 77.84 & 69.08 &  \\
% \midrule
% % FedSIM & & & &  & & & & & \\
% pFedMe & 78.10 & 72.60 & 79.59 & 70.88 &  \\
% FedLoRA& 77.64 & 73.57 & 79.53 & 70.58 & \\
% PERADA & 78.11  & 71.47 & 79.60 & 71.58 & \\
% FedSDR & 61.98 & 64.83 & 71.24 & 70.33 & \\
% \midrule
% FedOA & 77.39 & 72.22 & 79.12 & 70.92 & \\
% \bottomrule
% \end{tabular}
% \end{center}
% \end{table}

\subsection{Scalability Analysis}
To evaluate the scalability of our approach, we conducted experiments with an increased number of clients (up to 30) across four datasets, comparing our method to four personalized methods and one global model method. As shown in Table~\ref{table-scale}, our method consistently outperformed other personalized methods, demonstrating superior stability and effectiveness in expanded client scenarios. Furthermore, under more heterogeneous settings, FedOA exhibited greater stability than other personalized methods and achieved results comparable to the global model. These findings underscore the scalability of our approach, making it well-suited for larger and more complex federated settings while maintaining high performance.

\begin{table}[h]
\caption{Ablation study of scalability with 30 clients using ``leave-one-task-out'' validation. FedIT is tested on a single global model, while the remaining models are tested on personalized models with average results reported. Reading Com represents the Reading Comprehension task.  }
\label{table-scale}
\begin{center}
\begin{tabular}{l|llll|l}
 \toprule
\bf {Methods} &  {\bf Entailment} & {\bf Sentiment} &  {\bf Paraphrase} & {\bf Reading Com} & {\bf Average} \\
\midrule
\textcolor{gray}{FedIT} & \textcolor{gray}{41.50} &  \textcolor{gray}{78.75} &  \textcolor{gray}{44.50} &  \textcolor{gray}{58.04} & \textcolor{gray}{55.70} \\
\midrule
% FedSIM & & & &  & & & & & \\
pFedMe & 31.44 & 61.36 & 37.75 & 38.52 & 42.27 \\
FedLoRA& 34.89 & 65.49 & 36.74 & 46.90 & 46.01 \\
PERADA & 31.41  & 61.33 & 37.67 & 39.30 & 42.43 \\
FedSDR & 29.19 & 42.79 & 32.94 & 26.95 & 32.97 \\
\midrule
FedOA & \textbf{38.99} & \textbf{78.33} & \textbf{44.65} & \textbf{58.84} & \textbf{55.20}\\
\bottomrule
\end{tabular}
\end{center}
\end{table}

\subsection{Adaptability Analysis}

To enhance applicability across diverse non-IID environments, our method is strategically designed for high flexibility, enabling adaptation across various global learning frameworks, backbones and PEFT methods for different scenarios. This adaptability is simply achieved through the straightforward substitution of the FedAvg, LLM and LoRA with alternative aggregation methods, transformer-based foundation models and adapter-based PEFT methods during the training. In our experiment, we employ FedAvg, LLM and LoRA as representative examples, demonstrating our methods' superior performance compared to other baselines as indicated in Table~\ref{table2}.

To further validate the effectiveness and versatility of our approach across different federated foundation model contexts, we extend our methods to include the ViT \cite{dosovitskiy2020image} framework and also implement other baselines within ViT to maintain a fair comparison. We conduct experiments on OfficeHome datatset \cite{venkateswara2017deep},which comprises images across four distinct domains with 65 categories. In line with our previous experiments, we employed a "leave-one-domain-out" strategy, where each of the three clients maintains data from one distinct domain, setting aside the remaining domain as the testing data for evaluating OOD generalization.
Results presented in Table~\ref{table-vit} indicate that our methods outperform other personalized models and have comparable results with global models. These findings underscore the robustness and consistent efficacy of our methods across various federated foundation models context. 

\begin{table}[h]
\caption{OOD results of different models using ``leave-one-domain-out'' validation. FedIT is tested on a single global model, while the remaining models are tested on personalized models with average results reported.}
\label{table-vit}
\begin{center}
\begin{tabular}{l|llll|l}
 \toprule
\bf {Methods} &  {\bf Art} & {\bf CliPart} &  {\bf Product} & {\bf Real World} & {\bf Average} \\
\midrule
\textcolor{gray}{FedIT} & \textcolor{gray}{68.11} & \textcolor{gray}{56.66} & \textcolor{gray}{77.18} & \textcolor{gray}{77.94} & \textcolor{gray}{69.97}\\
\midrule
% FedSIM & & & &  & & & & & \\
pFedMe & 54.72 & 41.25 & 59.22 & 60.67 & 53.96 \\
FedLoRA& 60.49 & 51.31 & 72.93 & 73.15 & 64.47 \\
PERADA & 54.73  & 41.25 & 59.24 & 60.68 & 53.98\\
\midrule
FedOA & \textbf{67.49} & \textbf{56.51} & \textbf{75.96} & \textbf{77.45} & \textbf{69.35}\\
\bottomrule
\end{tabular}
\end{center}
\end{table}

%%%%%%%%%%%%%%%%%%%%%%%%%%%%%%%%%%%%%%%%%%%%%%%%%%%%%%%%%%%%

\newpage

\end{document}